\pdfoutput=1
\documentclass[11pt]{article}
\usepackage[section]{placeins}
\usepackage{listings}
\usepackage{xcolor}

\usepackage{subcaption}
\usepackage{graphicx}
\usepackage{enumitem}
\usepackage[final]{acl}

\usepackage{times}
\usepackage{latexsym}
\usepackage[inkscapelatex=false]{svg}
\usepackage{ragged2e}
\usepackage[T1]{fontenc}

\usepackage[utf8]{inputenc}
\usepackage{booktabs}
\usepackage{microtype}

\usepackage{inconsolata}
\usepackage{tabularx}
\usepackage{algorithm}
\usepackage{algpseudocode}
\usepackage{amsmath}
\usepackage{bbold}
\usepackage{natbib}
\usepackage[normalem]{ulem}
\definecolor{darkpastelgreen}{rgb}{0.01, 0.75, 0.24}
\definecolor{emerald}{rgb}{0.31, 0.78, 0.47}
\algrenewcommand\algorithmicindent{0.8em}
\algrenewcommand{\algorithmiccomment}[1]{%
  \hfill{\footnotesize$\triangleright$~\textit{#1}}}

\title{\textsc{MultAttnAttrib}: Training-Free Multimodal Attribution \\in Long Document Question Answering}

\author{
  \textbf{Dang Quang Thien Tran}$^{1}$\thanks{\ \,Equal contribution.}
  \quad \textbf{Quang V. Dang}$^{1*}$
  \quad \textbf{Vinamra Tyagi}$^{1}$ \\
  \textbf{Sai Soorya Rao Veeravalli}$^{1}$
  \quad \textbf{Trang Nguyen}$^{1}$
  \quad \textbf{Ryan A. Rossi}$^{2}$ \\
  \textbf{Franck Dernoncourt}$^{2}$
  \quad \textbf{Nedim Lipka}$^{2}$
  \quad \textbf{Koustava Goswami}$^{2}$
  \quad \textbf{Samyadeep Basu}$^{2}$ \\[0.4em]
  $^{1}$University of Massachusetts, Amherst \qquad $^{2}$Adobe Research, San Jose
}

\begin{document}
\maketitle

\begin{abstract}

As grounded QA systems are increasingly deployed in AI assistants, accurately attributing generated answers to evidence is critical for user trust and model safety. While unimodal attribution has been explored in depth, the multimodal setting remains under-researched. We introduce \textsc{MultAttnAttrib}\footnote{\href{https://github.com/dqttran/MultAttnAttrib}{\nolinkurl{github.com/dqttran/MultAttnAttrib}}}, a training-free and label-supervised attribution method that reads a model's prefill pass, selected attention heads, and calibrated thresholds to locate source evidence within a document. Across four VLMs, we find that retrieval heads are broadly shared across modalities, but the highest-ranked retrieval heads are modality-specific, with image-retrieval heads consistently concentrated in deeper layers than text-retrieval heads. Given the lack of fine-grained multimodal attribution in long-form documents, we also propose \textsc{MultAttrEval}\footnote{\href{https://huggingface.co/datasets/dqttran/MultAttrEval}{\nolinkurl{huggingface.co/datasets/dqttran/MultAttrEval}}}, which comes with validated ground-truth attributions across six domains. \textsc{MultAttnAttrib} outperforms every prompting and RAG configuration across all three regimes on Qwen3-VL while requiring one-seventh of the direct-prompting inference latency, and remains competitive with frontier models such as GPT-5.4.

\end{abstract}

\section{Introduction}
Building user trust in AI systems is critical to the success of agentic workflows in both enterprise and consumer environments. In many settings, users cannot safely act on a generated answer without verifying its source and validity — even modern generative systems fully support fewer than 52\% of their generated statements with accurate citations \citep{liu2023evaluating}. As a result, model grounding via \textbf{\textit{attributions}}—localizing each answer component to its supporting evidence—has emerged as a fundamental requirement for model deployment, particularly in domains such as medicine where ungrounded or hallucinated answers can have real negative impacts \citep{kim2025medical}.

There have been increasing efforts to use attributions to ground Document Question-Answer pairs, though these efforts largely focused on unimodal contexts. Currently, unimodal attributions are often generated using citation-style generation \citep{bohnet2022attributed, gao2023enabling, berchansky2024cotar}, or model circuit/retrieval head isolation ~\citep{basu2025mechanistic}. Another line of research involves attributing generated text by decomposing it into smaller units~\citep{ramu2024enhancing,balasubramanian-etal-2026-decomposition}, but it has only explored text-only attribution. Real documents, however, routinely interleave text with images, charts, and other raster content. A robust attribution system must identify not only the correct source but also the supporting modality, and arbitrate when evidence spans both.

Subsequently, the multimodal long-document setting remains comparatively nascent, with preexisting approaches largely framing attribution as citation selection from pre-retrieved passages or images rather than as fine-grained localization within a single full-length document~\citep{ma2025visa, Qi_2024, song2026mavis}. A unique set of challenges arises in multimodal attribution that doesn't apply in the unimodal case: determining both the correct modality (or modalities) and the correct source within it. Resolving how text and images jointly support a single attribution remains an open problem with significant downstream potential.

To address this, we propose \textbf{\textsc{MultAttnAttrib}}, a parameter-training-free multimodal attribution method that uses a small set of labeled data to identify retrieval heads and leverage them for direct attribution in one pass. The backbone remains frozen throughout; that is, no gradient updates or fine-tuning are performed, and no auxiliary module is trained. Instead, a one-time offline procedure uses 90 labeled items to select eight existing attention heads and calibrate two modality thresholds. At inference, the method aggregates attention from these heads to score text spans and whole images jointly, producing modality-aware citations in a single pass.

Current benchmarks are narrowly scoped, rely on pre-retrieved candidate pools, or are not designed for multimodal attribution. We therefore pair \textsc{MultAttnAttrib} with \textbf{\textsc{MultAttrEval}}, spanning six domains with unimodal and multimodal settings, and evaluate prompting and post-hoc attribution baselines. \textsc{MultAttnAttrib} is the only method evaluated above $0.58$ F1 in all three regimes on Qwen3-VL~\citep{bai2025qwen3} and outperforms GPT-5.4~\citep{openai2026gpt54} on visual grounding. Relative to direct Qwen3-VL prompting, it runs at $\sim14\%$ of the latency and uses $\sim15$\,GB less peak memory per QAA instance.

In summary, our contributions are as follows.
\begin{itemize}[nosep]
\item \textbf{\textsc{MultAttnAttrib}}: A training-free and label-supervised multimodal attribution method that produces modality-aware citations efficiently in a single inference pass.
\item \textbf{\textsc{MultAttrEval}}: A complementary benchmark for fine-grained multimodal attribution in long documents across six domains.
\item Extensive experiments showing that \textsc{MultAttnAttrib} outperforms every prompting and RAG configuration on Qwen3-VL across all three regimes at substantially lower latency, while generally outperforming VLM prompting across three additional backbones.
\item An analysis across four backbones reveals that retrieval heads are broadly shared across modalities, but top-ranked heads are modality-specific and image retrieval heads consistently emerge in deeper layers.
\end{itemize}

\section{Related Work}
\label{sec:related}
\subsection{Attribution on Multimodal Inputs}
The unreliability of language model outputs has motivated extensive work on citing and attributing generated text, falling into three broad families. The first family fine-tunes models to interleave citations with output, building on Attributed QA \citep{bohnet2022attributed}, the ALCE benchmark \citep{gao2023enabling}, and training-based citation generation methods \citep{aly2024learning, asai2024self, huang2024training, li2024improving, slobodkin2024attribute, zhang2025longcite, chuang2025selfcite}. These methods operate on text-only inputs and so cannot produce image or multimodal attributions for interleaved documents. The second family decouples attribution from generation by post-processing outputs with external retrievers, NLI verifiers, or LLM judges \citep{gao2023rarr, qian2025vericite}. The third family recovers attribution directly from the model's computations: by aggregating attention signals across heads \citep{basu2025mechanistic, ding2025attention, wang2025attntrace}, by reading internal signals via saliency maps or intermediate activations \citep{Qi_2024, phukan2024peering, phukan2025beyond}, or by probing the model through systematic context ablations \citep{cohen2024contextcite}. Our method belongs to this third family.

\subsection{Datasets for Multimodal Attribution}
Evaluating multimodal attribution requires benchmarks that test evidence localization over full multimodal documents. Existing benchmarks such as MCiteBench~\citep{hu2025mcitebench}, MMDocRAG~\citep{dong2026benchmarking}, and MAVIS~\citep{song2026mavis} instead evaluate citation selection from small, pre-curated pools of passages, figures, or tables, reducing attribution to discrete candidate selection rather than true localization. Similarly, SciClaimEval~\citep{ho2026sciclaimeval} pre-identifies the relevant figure or table and evaluates only cross-modal entailment, sidestepping retrieval entirely. These settings do not reflect deployment conditions, where models must localize supporting evidence within long, interleaved documents. Concurrent work, MuRGAt~\citep{wan2026multimodal}, also studies free-form evidence selection without a candidate pool, but focuses on temporal video/audio attribution and generation-based methods. In contrast, our approach extracts citations directly from attention signals over static multimodal documents in a single forward pass.
\section{\textsc{MultAttnAttrib}: A Training-Free Approach for Multimodal Attribution}
\begin{figure*}[t]
    \centering
    \small
    \includegraphics[trim=0 18.5cm 0 1cm, width=\linewidth, clip]{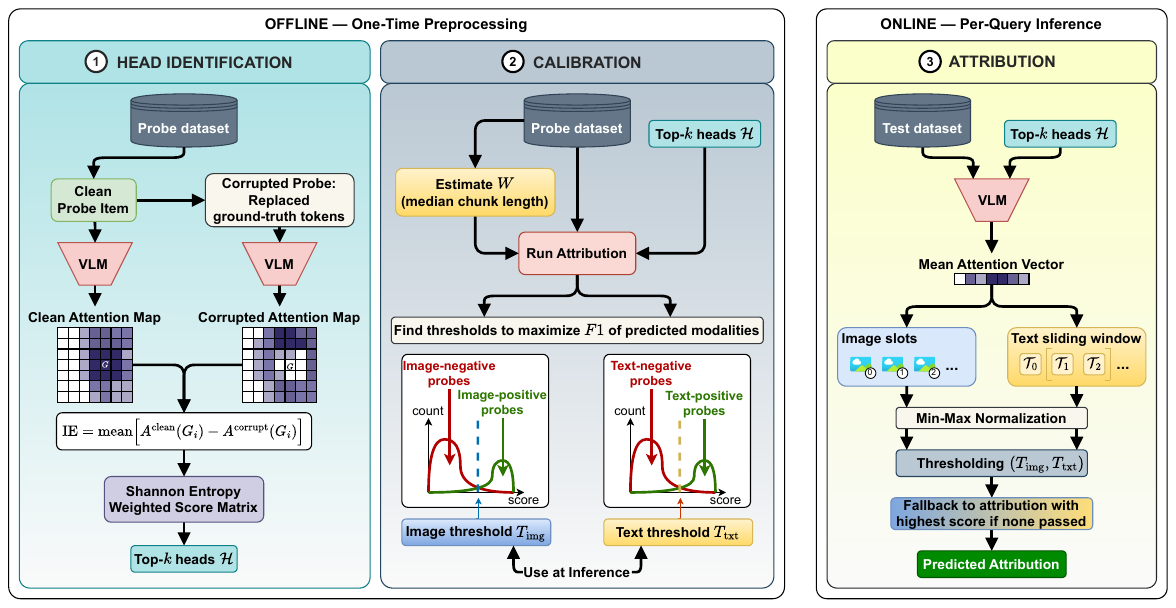}
    \caption{\textbf{\textsc{MultAttnAttrib}}: We identify signals for each attention head, then filter to select cross-modal heads. We then calibrate the threshold to maximize F1 scores on the probe set from \textsc{MultAttrEval}. For attribution, we use our top $k$ heads to generate attention spans and return the final results using our calibrated thresholds.}
    \label{fig:attn_diagram}
\end{figure*}

Mechanistic interpretability offers an alternative to high-cost model tuning/orchestration (Figure~\ref{fig:attn_diagram}). We identify a sparse subset of attention heads responsible for copying evidence from context, then attribute via the attention maps in a single forward pass \citep{basu2025mechanistic, wu2025retrieval}. However, these approaches focus on text-only QA, leaving image and multimodal QA unattributed. We find that retrieval heads are modality-specific at the top ranks but largely shared across the broader population. This motivates \textsc{MultAttnAttrib}, a label-supervised, training-free approach that exploits this shared backbone by identifying \textit{cross-modal} retrieval heads from a small probe set and extracting their attention signals to score image slots and text passages in a single forward pass. Here, training-free refers specifically to the absence of model-parameter updates: the backbone remains frozen, and no auxiliary model is trained. Supervision is limited to a small, 90-item labeled probe set used once offline to rank and select existing attention heads and calibrate two scalar modality thresholds.

The deviation from \citet{wu2025retrieval} is threefold. First, copy-paste frequency has no analog for image patches and identifies heads correlationally, whereas we score by causal mediation under evidence corruption. Section~\ref{sec:results} isolates the resulting visual evidence gap via MAS ablation. Second, the reported sparsity is a replicated finding rather than a novel claim. Third, the contribution lies in the \emph{allocation} of this sparse population across modalities, including a layer-wise separation between image and text retrieval that a text-only setting cannot expose (Section~\ref{sec:head_analysis}).

\subsection{Task}

We study the problem of \emph{multimodal attribution}. Given a set of documents composed of text and images, a question, and an answer, the goal is to correctly attribute the answer to its supporting evidence in the multimodal context provided.

Let the document be $\mathcal{D} = (\mathcal{T}, \mathcal{I})$, where $\mathcal{T} = (t_1, t_2, \dots, t_{|\mathcal{T}|})$ is a sequence of text tokens, and $\mathcal{I} = \{ I_1, I_2, \dots, I_{|\mathcal{I}|} \}$ is a set of images in the document. A text span is defined as a contiguous subsequence $\mathcal{T}_{i:j} = (t_i, t_{i+1}, \dots, t_j)$, with $1 \le i \le j \le |\mathcal{T}|$. Given a question $q$, the system produces an answer $a$, which is attributed to one of the following evidence types: a text span $\mathcal{T}_{i:j}$, a set of images ${\mathcal{I}}^* \subseteq \mathcal{I}$, or a joint pair. We define the attribution space as $\mathcal{A} = \bigl\{\, \mathcal{T}_{i:j} \,\bigr\} \cup \bigl\{\, {\mathcal{I}}^* \,\bigr\} \cup \bigl\{\, (\mathcal{T}_{i:j},\, {\mathcal{I}}^*) \,\bigr\}$. The task is to learn a function $f : (q, \mathcal{D}, a) \rightarrow \hat{\alpha}$, where $\hat{\alpha} \in \mathcal{A}$ is the predicted attribution.

\subsection{Head Identification} \label{sec:head_identification}

To identify both unimodal and multimodal retrieval heads, we compare two scoring methods: Causal Mediation Analysis (CMA) and Mean Attention Scoring (MAS). Prior approaches, such as average copy-paste frequency \citep{wu2025retrieval} and path patching \citep{basu2025mechanistic,wang2022interpretability}, are either correlational or prohibitively expensive. Both CMA and MAS score all $L \times H$ heads on a fixed probe set of 90 labeled multimodal QAA triplets (30 per regime) from \textsc{MultAttrEval}, disjoint from the 608-item test split, with the same probe set reused for threshold calibration (Section~\ref{sec:calibration}), so no test item enters either stage and no cross-validation is performed. Because CMA consistently outperforms MAS, \textsc{MultAttnAttrib} adopts CMA.

\paragraph{MAS requires only a single forward pass per probe.} MAS adapts the retrieval-head criterion of \citet{wu2025retrieval}, whose copy-paste frequency has no analog for image patches, to score heads on their attention instead. Heads are scored by how strongly they attend to a ground-truth token $G_i$ relative to a typical token elsewhere in the document $D_i$, measuring how selectively a head attends to evidence over distractors. We compare per token rather than in total to keep a short evidence span from being drowned out by the far larger remainder of the document. MAS is cheaper than CMA (below) but correlational: a head that happens to concentrate on the ground-truth region scores high whether or not it causally mediates retrieval.

\paragraph{Adapting CMA for retrieval head identification} Our method requires only two forward passes per probe: a clean pass on the original input $x_i$, and a corrupted pass where the evidence is replaced with content from another document. While prior CMA work focused on text \citep{basu2025mechanistic}, we extend this corruption strategy to the multimodal setting: text evidence is replaced with an equal-length contiguous span from another probe's document, while corrupted images are resized to match the ground-truth image dimensions, preserving the patch grid. Matching shapes across passes is essential. Since the evidence occupies identical positions in both, any drop in attention reflects dependence on evidence \emph{content}, not a fixed spatial prior (the failure mode correlational scoring cannot rule out).

The Indirect Effect (IE) for each head $(l, h)$ is the difference in mean attention to ground-truth positions $G_i$ between clean and corrupted inputs, averaged over query tokens $Q_i$ (answer and question tokens, minus stopwords/punctuation). To suppress heads that merely spread attention uniformly, we downweight scores using normalized Shannon entropy \citep{zhai2023stabilizing} over document-averaged clean attention \citep{clark2019does}. Scores are accumulated across all probes, and the top-$k$ heads $\mathcal{H}$ are selected as retrieval heads. Pseudocode for both scoring methods appears in Appendix~\ref{app:algorithms}.

\subsection{Calibration}
\label{sec:calibration}
Using the selected heads $\mathcal{H}$, we run Attribution (Algorithm~\ref{alg:attribution}) on all probes. Windowing applies to text only: we slide a window of length $W$, the median chunk token length in the probe set, across text positions, while each image is one slot scored by averaging over its patch tokens. We normalize the two scores onto a common $[0,1]$ scale. Raw scores are not probabilities, so we partition them by ground-truth modality label and sweep an F1-maximizing threshold over the per-item maximum to select $T_\mathrm{img}$ and $T_\mathrm{txt}$. These are fixed once offline and reused at inference. Pseudocode is in Algorithm~\ref{alg:calibration}
(Appendix~\ref{app:algorithms}).

\begin{algorithm}
\caption{\textsc{MultAttnAttrib}: Attribution}
\label{alg:attribution}
\begin{algorithmic}[1]
\small
\Require $g_\phi$ (language model), $x$ (input prompt), $Q$ (query position), $\mathcal{H}$ (selected heads), $W$ (span length), $\beta_\mathrm{img}$, $\beta_\mathrm{txt}$
\State $A \leftarrow g_\phi(x,\, Q,\, \mathcal{H})$
\State $\bar{a} \leftarrow \operatorname{mean}_{(l,h),q} A_{l,h,q}$
\State $v_s^\text{img} \leftarrow \bar{a}_s$ \textbf{for} each image slot $s$
\State $v_w^\text{txt} \leftarrow \bar{a}_w$ \textbf{for} each text window $w$
\State $[v^\text{img},\, v^\text{txt}] \leftarrow$ \textsc{MinMaxNorm}$([v^\text{img},\, v^\text{txt}])$
\State $\hat{\mathcal{I}}(\tau) \leftarrow \{s : \max v^\mathrm{img} \geq \tau,\; v_s^\mathrm{img} \geq \beta_\mathrm{img} \max v^\mathrm{img}\}$
\State $\hat{\mathcal{T}}(\tau) \leftarrow \{w : \max v^\mathrm{txt} \geq \tau,\; v_w^\mathrm{txt} \geq \beta_\mathrm{txt} \max v^\mathrm{txt}\}$
\If{$\hat{\mathcal{I}}(T_\mathrm{img}) \cup \hat{\mathcal{T}}(T_\mathrm{txt}) = \emptyset$}
    \State $T_\mathrm{img},\, T_\mathrm{txt} \leftarrow \max_m \max v^m$
\EndIf
\State \Return $\hat{\mathcal{I}}(T_\mathrm{img})\,,\, \hat{\mathcal{T}}(T_\mathrm{txt})$
\end{algorithmic}
\end{algorithm}

\subsection{Attribution}
Attribution requires a single forward pass over the query document. We average attention across the selected heads and score image slots and text windows as in Section~\ref{sec:calibration}, citing a modality only when its best score clears its threshold. We cite the top-scoring text window and every image scoring within a relative margin of the best (see \ref{sec:implementation}), since an answer can rest on more than one piece of evidence; if neither modality clears, we fall back to the highest-scoring one.

\section{\textsc{MultAttrEval}: A Dataset for Multimodal Attribution in Long Document Understanding}
\textsc{MultAttrEval} was created to address the need for fine-grained, domain-wide \textbf{\underline{Q}}uestion-\textbf{\underline{A}}nswer-\textbf{\underline{A}}ttribution (\textbf{QAA}) triplets given long documents with mixed-modality content. The dataset was generated in two waves, a first wave using 20 documents and a second using 29 additional documents.

\begin{figure*}[t]
    \centering
    \includegraphics[width=\linewidth, clip]{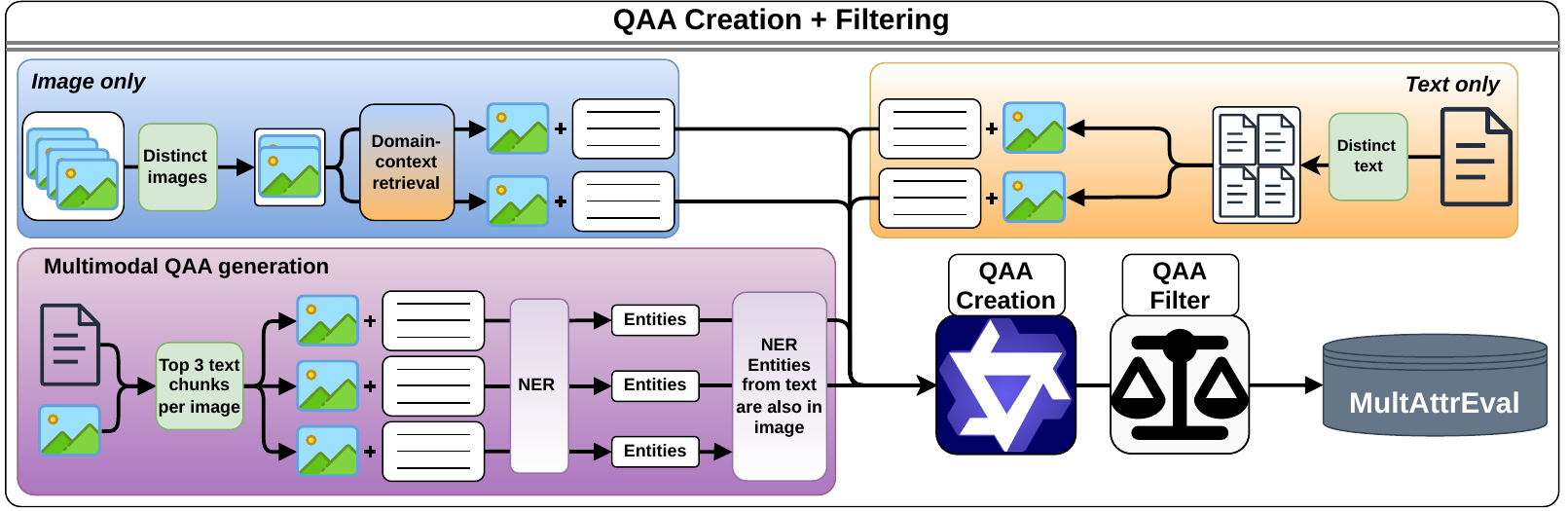}
    \caption{\textbf{\textsc{MultAttrEval}}: Overview of the QAA generation process used to construct MultAttrEval from processed MINT-1T PDFs across text-only, image-only, and combined text-image attribution settings. Candidate QAAs are model-generated, automatically filtered, and then human validated on a sampled subset.}
    \label{fig:data_gen_diagram}
\end{figure*}

Dataset construction proceeds in three stages. First, an MLLM generates candidate QAAs from selected source evidence. Second, modality-specific automated gates filter candidates for evidential support, non-triviality, and answerability. In the second-wave expansion, these gates retained 1,587 of 3,906 generated candidates (40.6\%). Third, human annotators evaluated a modality-stratified sample of 689 retained QAAs to estimate overall dataset quality; not every retained QAA was manually reviewed.

\subsection{Dataset Generation}

As shown in Figure~\ref{fig:data_gen_diagram}, we begin with MINT-1T PDFs \citep{awadalla2024mint}, filtered on image count and URL validity to ensure sufficient multimodal content. We semantically chunk the text, extract images from the TIFF frames, and project both
into a shared space via Cohere Embed v4.0~\citep{cohere2025embedv4}, so similarity is computed at the passage and image level rather than over whole documents or pages. QAAs are then generated \emph{backward from evidence}: an MLLM is conditioned on selected evidence, so every QA is tied to source content by construction and needs no post-hoc alignment.

\paragraph{\textbf{Unimodal (Text/Image only).}}
For text-only QAAs, we select diverse text chunks with low similarity to images and to other chunks, then generate questions and answers from each selected chunk alone. For image-only QAAs, we filter visually similar images and retrieve moderately related text chunks as non-attributed context for document domain and terminology. Generation and verification require every answer-specific claim to be supported by the raster rather than by this contextual text.

\paragraph{\textbf{Text + Image.}}
We select high-similarity image--text pairs, extract named entities from the text, and retain entities that are visually verified in the paired image. The surviving image, text, and verified entities are used to generate QAAs, which are subsequently filtered for QA quality and support from both modalities. Near-duplicate images are clustered to reduce redundant generation.

\begin{table}[h!]
    \centering
    \small
    \begin{tabularx}{\linewidth}{@{}X|c|ccc|c@{}}
        \toprule
        Domain & Docs & Text & Image & Both & \textbf{Total} \\
        \midrule
        Academic  & 13 & 135 & 205 & 252 & \textbf{592} \\
        Business  & 10 & 90  & 213 & 123 & \textbf{426} \\
        Finance   & 8  & 80  & 286 & 138 & \textbf{504} \\
        Health    & 9  & 86  & 141 & 201 & \textbf{428} \\
        Legal     & 6  & 52  & 95  & 87  & \textbf{234} \\
        Marketing & 3  & 23  & 33  & 45  & \textbf{101} \\
        \midrule
        \textbf{Total}     & \textbf{49} & \textbf{466} & \textbf{973} & \textbf{846} & \textbf{2285} \\
        \bottomrule
    \end{tabularx}
    \caption{\textsc{MultAttrEval} distribution across both waves, document domains, and modality regimes}
    \label{tab:multattreval_dist}
\end{table}

\subsection{Quality Verification}

The ground truth in \textsc{MultAttrEval} is constructive. Because QAs are generated \textit{from} evidence, the truth attribution is known by construction and needs only validation, not alignment.

\paragraph{\textbf{Automated filtering.}}
Each modality regime passes through its own agentic quality gate. Text-only items are checked for answer support, evidence-span validity, and cross-chunk uniqueness; image and multimodal items are additionally checked for source-independent wording, non-triviality, and answerability. Combined items must also ground distinct answer content in both modalities. Full acceptance criteria and rubrics appear in Appendix~\ref{app:multattreval_qaa} and Appendix~\ref{app:rubrics}.

\paragraph{\textbf{Human-in-the-loop.}} Using AI agents to generate \textit{and} verify data can provide dubious results. As a result, we included a human validation study over 689 samples of the filtered QAAs, sampled equally across modalities. Each item was independently judged by two of six annotators for answer correctness and for whether the ground-truth attribution supports the answer. Annotators judged 93.3\% of answers correct and 91.9\% of attributions supportive, with pairwise agreement of 90.4\% and 89.0\% respectively. Per-modality breakdowns and full agreement tables are in Appendix~\ref{app:annotation} and~\ref{app:multattreval_qaa}.

\textsc{MultAttrEval} contains 2,285 QAA triplets over 49 long-form PDF
documents in six domains, built in two waves: 20 documents (698 triplets) and an expansion of 29 (1,587). Items are 20.4\% text-only, 42.6\% image-only, and 37.0\% combined, the last being the share of answers whose ground truth spans both modalities (Table~\ref{tab:multattreval_dist}). Unless noted otherwise, \textbf{all experiments} use the first-wave split; the expansion wave is released as a held-out extension and used only as a consistency check, and specific details are offered in Appendix~\ref{app:expansion}.

\section{Experiments}
\subsection{Implementation Details}
\label{sec:implementation}
All methods are evaluated on the 608-item first-wave test split, which excludes the 90 probe items used for head selection and calibration (Section~\ref{sec:head_identification}). Every evaluated method receives the \emph{entire} document and must localize within it, so full-document difficulty is preserved. \textsc{MultAttnAttrib} uses the top $k = 8$ CMA-scored heads and a sliding window of $W = 144$ tokens; calibration yields $T_\mathrm{img} = 0.005$ and $T_\mathrm{txt} = 0.185$. The relative-margin parameters of Algorithm~\ref{alg:attribution} are set to $\beta_\mathrm{img} = 0.8$ and $\beta_\mathrm{txt} = 1.0$. At $\beta_\mathrm{img} = 0.8$, an answer resting on several comparable images cites all of them while long tails of weak matches are suppressed; at $\beta_\mathrm{txt} = 1.0$ only the top-scoring window clears the margin, so exactly one text span is cited per item. All runs use a single A100 at batch size 1 without vLLM.

\subsection{Baselines}
\paragraph{\textbf{Prompting.}} We evaluate \textsc{Qwen3-VL-30B-A3B-Instruct}~\citep{bai2025qwen3} and \textsc{GPT-5.4}~\citep{openai2026gpt54}. The \textbf{VLM} baseline uses images and OCR text, while the \textbf{LLM} baseline replaces images with pure text captions. We also evaluate \textbf{RAG} variants retrieving five sources per query: \texttt{Cohere} retrieves text chunks and images independently, and \texttt{ColQwen}~\citep{huang2025beyond} retrieves full PDF pages.

\paragraph{\textbf{Other Methods.}} We compare against \textbf{MAVIS}'s included method \citep{song2026mavis}, \textbf{TracLLM} \citep{TracLLM}, \textbf{M3DocRAG} \citep{cho2024m3docrag}, and a \textbf{MAS} variant of our method (Section~\ref{sec:head_identification}), representing various approaches discussed in Section~\ref{sec:related}.

\paragraph{\textbf{MAVIS.}} We adapt MAVIS from candidate selection to localization by numbering all text blocks and figures and prompting sentence-final \texttt{[i]} citations over the full document. Unlike the other baselines, MAVIS natively cites images. Text units use native paragraph chunks approximating \textsc{MultAttrEval}'s gold span granularity (Appendix~\ref{app:baseline}).

\subsection{Evaluation Metrics}

We evaluate attribution quality using macro-averaged precision, recall, and $\mathrm{F1}$. Images use exact ID matching, while text spans use a three-tiered fuzzy substring similarity and penalties for under- or over-quoting by length ratio. Because gold spans are exact substrings, we omit NLI-based entailment matching, avoiding added variance without improving discrimination. We score each predicted citation independently against the ground truth, which coincides with \textsc{MultAttrEval}'s one-span, one-image-set design (see Algorithm~\ref{alg:attribution} and Section~\ref{sec:implementation}, and diverges only on the multi-span benchmarks of Section~\ref{sec:other_datasets}. Full evaluation details and additional LLM-as-Judge are provided in Appendix~\ref{app:eval_metrics} and ~\ref{app:llm_judge}.

\section{Results and Analysis}
\label{sec:results}
\subsection{Comparing \textsc{MultAttnAttrib} on \textsc{MultAttrEval} with the Backbone and Other Attribution Methods}
\begin{table*}[t]
\centering
\small
\setlength{\tabcolsep}{3pt}
\begin{tabular*}{\linewidth}{@{\extracolsep{\fill}}lccccccccc@{}}
\toprule
 & \multicolumn{3}{c}{Text-only} & \multicolumn{3}{c}{Image-only} & \multicolumn{3}{c}{Text + Image} \\
\cmidrule(lr){2-4} \cmidrule(lr){5-7} \cmidrule(lr){8-10}
Method & P & R & F1 & P & R & F1 & P & R & F1 \\
\midrule
\multicolumn{10}{l}{Backbone: \textit{Qwen3-VL-30B-A3B-Instruct}} \\
\midrule
VLM & 0.666 & 0.382 & 0.485 & 0.477 & \textbf{0.871} & 0.617 & 0.524 & 0.465 & 0.493 \\
LLM & 0.696 & 0.402 & 0.510 & 0.403 & 0.813 & 0.539 & 0.552 & 0.396 & 0.461 \\
Cohere + VLM & \textbf{0.883} & 0.380 & 0.531 & 0.427 & 0.813 & 0.560 & 0.713 & 0.453 & 0.554 \\
Cohere + LLM & 0.877 & 0.415 & 0.563 & 0.466 & 0.858 & 0.604 & 0.725 & 0.459 & 0.562 \\
ColQwen + VLM & 0.721 & 0.376 & 0.494 & 0.400 & 0.698 & 0.508 & 0.581 & 0.484 & 0.528 \\
\midrule
MAVIS & 0.736 & 0.795 & \textbf{0.764} & 0.570 & 0.636 & 0.601 & \textbf{0.785} & 0.417 & 0.544 \\
TracLLM (STC) & 0.609 & \textbf{0.856} & 0.712 & 0.253 & 0.707 & 0.373 & 0.602 & 0.491 & 0.541 \\
M3DocRAG & 0.535 & 0.405 & 0.461 & 0.323 & 0.604 & 0.421 & 0.524 & 0.509 & 0.516 \\
\midrule
\multicolumn{10}{l}{\textit{\textsc{MultAttnAttrib} (Ours)}} \\
\midrule
Full Document (CMA) & 0.572 & 0.621 & 0.596 & \textbf{0.750} & 0.804 & 0.776 & 0.626 & 0.544 & 0.582 \\
MAS & 0.607 & 0.583 & 0.595 & 0.680 & 0.747 & 0.712 & 0.602 & 0.460 & 0.521 \\
Cohere (CMA) & 0.614 & 0.726 & 0.665 & 0.749 & 0.827 & \textbf{0.786} & 0.643 & \textbf{0.564} & \textbf{0.601} \\
ColQwen (CMA) & 0.609 & 0.619 & 0.614 & 0.605 & 0.662 & 0.632 & 0.628 & 0.541 & 0.581 \\
\midrule
\multicolumn{10}{l}{Backbone: \textit{InternVL3.5-30B-A3B}} \\
\midrule
VLM & \textbf{0.585} & 0.303 & 0.399 & \textbf{0.653} & \textbf{0.782} & \textbf{0.712} & 0.480 & 0.339 & 0.398 \\
Full Document (CMA) & 0.505 & \textbf{0.638} & \textbf{0.564} & 0.637 & 0.769 & 0.697 & \textbf{0.553} & \textbf{0.508} & \textbf{0.530} \\
\midrule
\multicolumn{10}{l}{Backbone: \textit{Mistral-Small-3.2-24B}} \\
\midrule
VLM & \textbf{0.720} & 0.325 & \textbf{0.448} & 0.592 & 0.653 & 0.621 & \textbf{0.571} & 0.287 & 0.382 \\
Full Document (CMA) & 0.327 & \textbf{0.615} & 0.427 & \textbf{0.726} & \textbf{0.787} & \textbf{0.755} & 0.524 & \textbf{0.530} & \textbf{0.527} \\
\midrule
\multicolumn{10}{l}{Backbone: \textit{Gemma-4-26B-A4B}} \\
\midrule
VLM & \textbf{0.837} & 0.294 & 0.435 & 0.398 & 0.502 & 0.444 & \textbf{0.721} & 0.295 & 0.419 \\
Full Document (CMA) & 0.571 & \textbf{0.662} & \textbf{0.613} & \textbf{0.746} & \textbf{0.813} & \textbf{0.778} & 0.552 & \textbf{0.490} & \textbf{0.519} \\
\bottomrule
\end{tabular*}
\caption{\textbf{\textsc{MultAttrEval} metrics (608-item
test split).} Existing MLLMs and attribution methods find visual evidence localization difficult, as evidenced by F1 scoring across modalities. \textsc{MultAttnAttrib} is the only method \emph{here} above 0.58 F1 in every regime on Qwen3-VL. The lower three blocks repeat the comparison between VLM prompting and our method on three more backbones, with the same 90-item probe and $k = 8$ heads. Bold marks the best value per column within each backbone block.}
\label{tab:main_results}
\end{table*}

\begin{figure*}[t]
    \centering
    \includegraphics[trim=0.5cm 1.0cm 0.5cm 0cm, width=0.95\linewidth, clip]{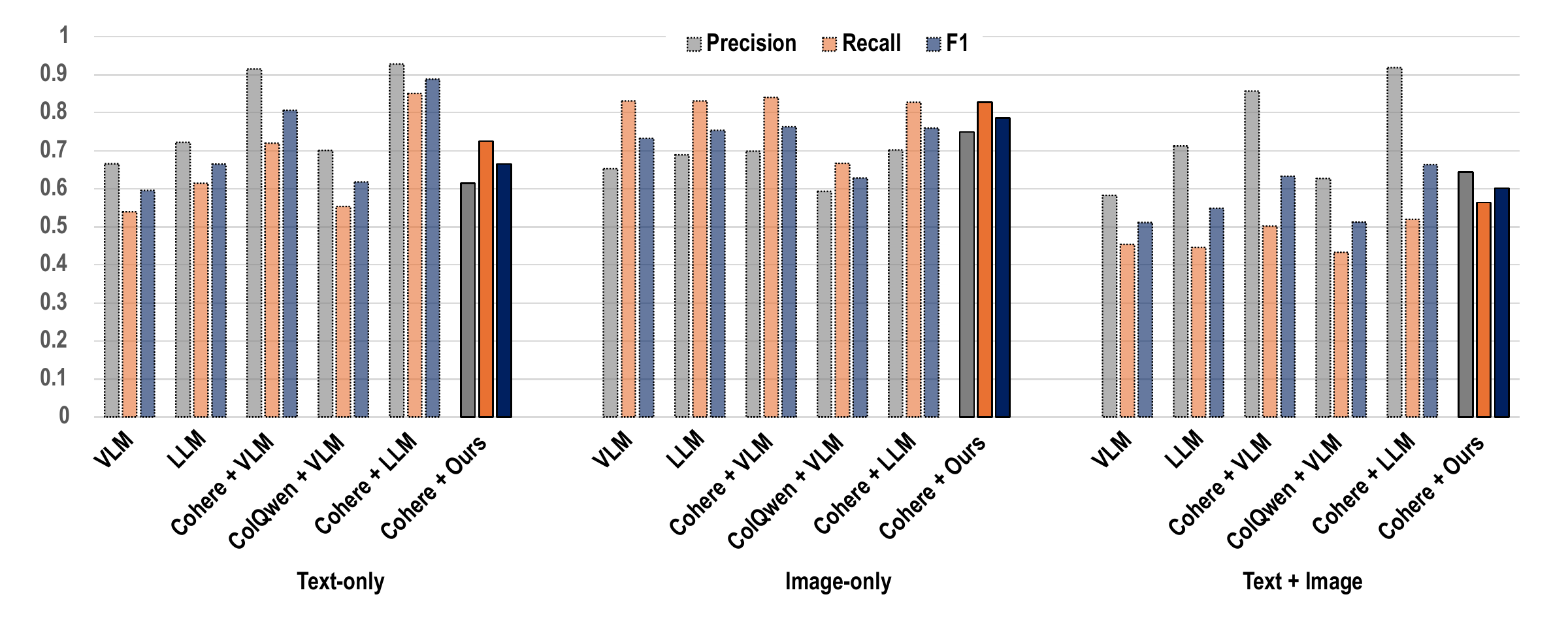}
    \caption{\textbf{\textsc{MultAttnAttrib} is competitive with GPT-5.4, a frontier closed-source model.} Comparing all GPT variants with the Cohere + \textsc{MultAttnAttrib} (Ours) variant.}
    \label{fig:gpt_comparison_chart}
\end{figure*}

\noindent\textbf{Scope.}
Unless otherwise stated, all analyses in this section use Qwen3-VL-30B-A3B-Instruct. Existing attribution methods are adapted to full-document localization to match our setup (Appendix~\ref{app:baseline}), and cross-backbone results are discussed separately.

As shown in Table~\ref{tab:main_results}, within the controlled Qwen comparison, \textsc{MultAttnAttrib} outperforms every prompting configuration in all three regimes. Relative to the strongest prompting result in each regime, the full-document variant gains $+0.033$ F1 on text-only, $+0.159$ on image-only, and $+0.020$ on combined attribution. With Cohere retrieval, these margins widen to $+0.102$, $+0.169$, and $+0.039$, respectively. These results establish the advantage of the attribution procedure under a matched backbone; we evaluate its transfer across backbones separately below.

\paragraph{\textbf{Probe supervision is lightweight and transferable.}}
On Qwen, repeating the full offline procedure with 30- and 60-item probe sets yields text/image/combined F1 of $.609/.773/.577$ and $.605/.779/.572$, respectively, compared with $.596/.776/.582$ for the 90-item setting. Thus, with a 30-item probe, F1 changes by at most $1.3$ points in any regime. Moreover, heads and thresholds derived from MCiteBench transfer to \textsc{MultAttrEval} without recalibration, achieving $.654/.706/.526$ and still outperforming matched Qwen VLM prompting in all three regimes. A Qasper-derived probe similarly achieves $.680$ text F1 (setup in Appendix~\ref{app:algorithms}).

{\paragraph{\textbf{The gains carry to other backbones.}} We repeat head identification and calibration on InternVL3.5-30B~\citep{wang2025internvl3}, Mistral-Small-3.2-24B~\citep{mistral2025small32}, and Gemma-4-26B~\citep{team2026gemma}, using the same 90-item probe. We keep $k = 8$ heads, the value set on Qwen3-VL, without per-model tuning. As the lower blocks of Table~\ref{tab:main_results} show, \textsc{MultAttnAttrib} outperforms VLM prompting on combined attribution on every backbone, by $0.100$ to $0.145$ F1. It also wins in most single-modality regimes. In the remaining two it trails by at most $0.021$. Setup details are in Appendix~\ref{app:backbones}.\par}

\paragraph{\textbf{Existing attribution methods lead on text but not across modalities.}} MAVIS leads text-only F1, exceeding our full-document variant by $0.168$, with TracLLM second. That advantage does not extend to visual evidence: TracLLM does not natively score image evidence, and MAVIS cites images but still trails us by $0.175$ F1 on image-only attribution. Thus, the visual-evidence advantage is not explained solely by whether a method can emit image citations; how that evidence is scored also matters. Within this comparison, \textsc{MultAttnAttrib} is the only method exceeding $0.58$ F1 in all three regimes. The MAS variant nearly matches CMA on text-only but trails it by $0.064$ F1 on image-only and $0.061$ on combined attribution, showing that causal head scoring matters most for cross-modal attribution.

\paragraph{\textbf{\textsc{MultAttnAttrib} responds more positively to fine-grained retrieval methods than page-based retrieval.}}Within Qwen experiments, combining our method with Cohere RAG substantially improves text performance and modestly improves image and multimodal results, while ColQwen degrades image metrics with only marginal multimodal changes. This suggests fine-grained retrieval is more effective than page-level retrieval.
\begin{figure*}[t]
  \centering
  \begin{subfigure}{0.35\linewidth}
    \centering
    \includegraphics[width=\linewidth]{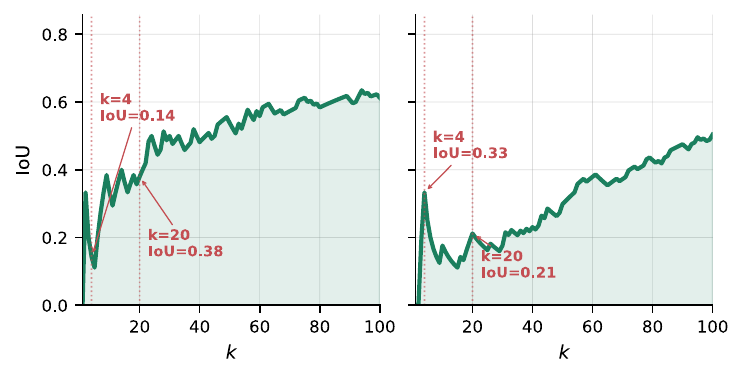}
    \caption{Top-$k$ head overlap (IoU)}
    \label{fig:iou}
  \end{subfigure}
  \hfill
  \begin{subfigure}{0.35\linewidth}
    \centering
    \includegraphics[width=\linewidth]{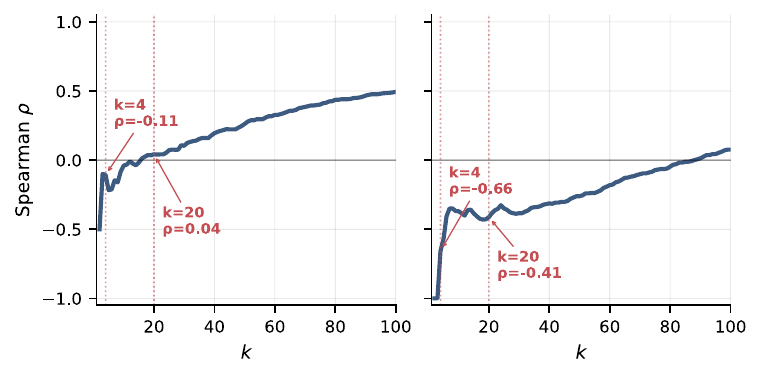}
    \caption{Spearman $\rho$ over the top-$k$ union}
    \label{fig:spearman}
  \end{subfigure}
  \hfill
  \begin{subfigure}{0.27\linewidth}
    \centering
    \includegraphics[width=\linewidth]{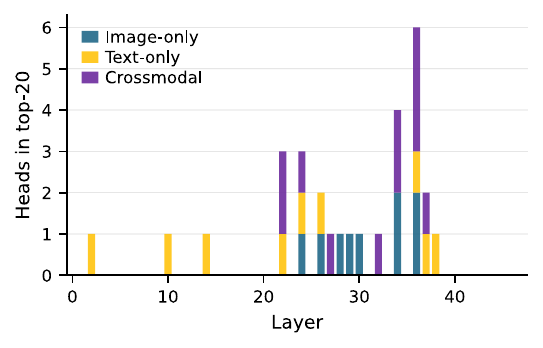}
    \caption{Layer distribution, CMA top~20}
    \label{fig:layer_counts}
  \end{subfigure}
  \caption{\textbf{Crossmodal retrieval head structure.} (a--b) CMA (left
  panel of each pair) vs.\ MAS (right): CMA yields higher
  image--text head overlap, and the broader head population is largely
  crossmodal with specialization only at the top ranks. (c) Image heads
  concentrate at mid-to-late layers while text heads span early to late
  layers; crossmodal heads cluster in the transition zone.}
  \label{fig:head_agreement}
\end{figure*}

\paragraph{\textbf{\textsc{MultAttnAttrib} processes long documents with lower memory and latency than VLM prompting.}}
Direct non-vLLM inference on our Qwen model results in frequent OOM errors for QA inputs, a problem \textsc{MultAttnAttrib} avoids by attributing in a single forward pass, bypassing KV-cache growth, and removing token-level decoding overhead. Focusing on the non-OOM QA pairings, \textsc{MultAttnAttrib} reduces peak VRAM from $78.28$ GB to $63.41$ GB and latency from $15.67 \pm 14.38$s to $2.16 \pm 0.17$s (a $7.3\times$ speedup) on a single A100 without vLLM at batch size 1.

\subsection{Comparing \textsc{MultAttnAttrib} on \textsc{MultAttrEval} to Frontier Prompting}
\label{sec:attnattrib_vs_gpt}
\textsc{MultAttnAttrib} shows complementary strengths relative to prompted GPT baselines (Figure~\ref{fig:gpt_comparison_chart}). On visual grounding, it achieves stronger image precision and F1 than all GPT baselines, since attention aggregation operates directly over the full token sequence rather than verbalizing attributions. In text settings, the trade-off shifts: GPT baselines attain higher precision by returning minimal evidence. At the same time, \textsc{MultAttnAttrib} recovers higher recall by capturing all influential tokens at the cost of including loosely related text. Overall, it remains competitive with frontier-scale closed-source models such as GPT-5.4 on multimodal attribution.

\subsection{Our Method on External Benchmarks}
\label{sec:other_datasets}

To test generalization beyond \textsc{MultAttrEval}, we evaluate on MCiteBench \citep{hu2025mcitebench} and Qasper \citep{dasigi2021qasper}. Both benchmarks' native tasks generate answers and evidence jointly, whereas \textsc{MultAttnAttrib} is post-hoc, so we supply each with an answer to attribute. For MCiteBench we regenerate answers from the ground-truth evidence, shared by all methods, because its answers leak evidence identifiers. For Qasper we use the released annotator answers. Scoring follows Section~\ref{sec:results}, extended to multi-span text (Appendix~\ref{app:eval_metrics}). \textsc{MultAttnAttrib} outperforms the VLM baseline in every regime (Table~\ref{tab:posthoc_external}), most on MCiteBench image-only attribution ($+0.197$ F1) and by $+0.053$ F1 on text-only Qasper. Under MCiteBench's native protocol, it trails GPT-4o on multi-source but leads on single-source explanation and locating (Appendix~\ref{app:mcite}).

\paragraph{Relaxing selection margin restores multi-span recall.} \textsc{MultAttrEval} answers rest on one passage, so there we cite only the top-scoring window ($\beta_\mathrm{txt} = 1.0$). Qasper and MCiteBench often need several, which is why we use $\beta_\mathrm{txt} = 0.8$ for them, motivated by the $\beta_\mathrm{img}$ used above. On Qasper, where 43.5\% of items have multiple gold spans, precision still rises 13.2 points and F1 5.3 points over the baseline. Lowering $\beta_\mathrm{txt}$ recovers more of the missing passages. From $1.0$ to $0.8$ to $0.5$, multi-span F1 rises on Qasper ($0.402 \rightarrow 0.455 \rightarrow 0.546$) and MCiteBench ($0.569 \rightarrow 0.604 \rightarrow 0.645$) without lowering single-span F1.

\begin{table}[h!]
\centering
\small
\setlength{\tabcolsep}{4pt}
\begin{tabularx}{\linewidth}{@{}Xl|ccc@{}}
\toprule
Benchmark  & Method & P & R & F1 \\
\midrule
MCiteBench, Text & VLM & 0.606 & 0.642 & 0.623 \\
                 (n = 1199) & Ours & \textbf{0.703} & \textbf{0.859} & \textbf{0.773} \\
                  \midrule
MCiteBench, Image & VLM  & 0.349 & \textbf{0.789} & 0.484 \\
                 (n = 1444) & Ours & \textbf{0.637} & 0.731 & \textbf{0.681} \\
                  \midrule
MCiteBench, T+I & VLM  & 0.690 & \textbf{0.668} & 0.679 \\
                (n = 236) & Ours & \textbf{0.737} & 0.637 & \textbf{0.683} \\
                  \midrule
Qasper, Text & VLM  & 0.481 & \textbf{0.521} & 0.501 \\
                 (n = 927) & Ours & \textbf{0.613} & 0.506 & \textbf{0.554} \\
\bottomrule
\end{tabularx}
\caption{Post-hoc attribution on two external benchmarks, with $\beta_\mathrm{txt} = \beta_\mathrm{img} = 0.8$.}
\label{tab:posthoc_external}
\end{table}
\subsection{Analysis of Unimodal and Crossmodal Attention Heads}
\label{sec:head_analysis}

A central design question for \textsc{MultAttnAttrib} is whether text and image retrieval share attention circuits or require modality-specific ones. We score all $L \times H = 1536$ heads under CMA and MAS, measuring cross-modal agreement via IoU and Spearman's $\rho$ over top-$k$ sets (Appendix~\ref{app:head_analysis}).

\paragraph{The head population is largely shared, but top-ranked heads are modality-specific.} A top text head (e.g., $(19,3)$) can be the worst image head. At $k=4$, both scorings are anti-correlated ($\rho_4 = -0.107$ CMA, $-0.657$ MAS), but CMA recovers by $k=20$ ($\rho_{20} = 0.042$) while MAS remains anti-correlated until $k=88$. CMA's overlap increases with $k$ ($\text{IoU}$ $0.143 \rightarrow 0.379$), whereas MAS's decreases ($0.333 \rightarrow 0.212$), suggesting CMA captures shared retrieval heads while MAS favors modality-specific attention.

\paragraph{Image retrieval is a late-layer phenomenon, but text retrieval is not.} Image heads concentrate in layers $22$--$36$ (Figure~\ref{fig:layer_counts}) while text heads span early and late layers, reflecting the richer syntactic and semantic demands of textual evidence; crossmodal heads cluster in the mid-to-late transition zone, forming the retrieval backbone common to both.

\paragraph{The retrieval circuit is extremely sparse.} About $80\%$ of heads score below $0.1$ after min-max normalization and fewer than $2\%$ exceed $0.6$ in any category, extending to the multimodal setting the finding that fewer than $5\%$ of heads qualify as retrieval heads \citep{wu2025retrieval} (Figure~\ref{fig:score_dist}).

\paragraph{The findings above replicate across three other models.} On InternVL3.5-30B-A3B, Mistral-Small-3.2-24B, and Gemma-4-26B-A4B, the top-ranked heads are modality-specific while overlap grows with $k$, image heads again start deeper than text heads, and only a small number of heads scores high (Appendix~\ref{app:backbones}).

\section{Conclusion}
We introduce \textsc{MultAttnAttrib}, a training-free attribution method with specialized cross-modal retrieval heads, which outperforms prompting strategies on the same backbone at a fraction of the latency across all three modality regimes, including visual evidence, where text-first methods degrade, and remains competitive with frontier-scale models such as GPT-5.4. Beyond Qwen3-VL, the same procedure, without per-model tuning, generally outperforms prompting on three further open-weight VLMs. We also introduce \textsc{MultAttrEval}, a testbed for multimodal attribution over long-form, multi-domain documents designed to reflect enterprise settings.

\section{Limitations}
\label{sec:limitations}

First, \textsc{MultAttnAttrib} reads attention from the prefill pass, so it requires white-box access to the backbone and cannot be applied to closed-weight API models, a constraint shared by internal-computation attribution generally. In exchange, it avoids token-level decoding, which yields the latency and memory reductions we report.

Second, the method is training-free but uses a small labeled probe set once offline to select heads and set thresholds. Reducing the probe from 90 to 30 items changes F1 by at most $1.3$ points, and probes derived from external benchmarks transfer without recalibration (Section~\ref{sec:results}, Appendix~\ref{app:algorithms}), so this dependence is limited.

Third, the text margin $\beta_\mathrm{txt}$ is set by hand to the expected number of gold spans, $1.0$ for \textsc{MultAttrEval} and $0.8$ for Qasper and MCiteBench. Lowering it raises multi-span F1 without hurting single-span items (Section~\ref{sec:other_datasets}), but choosing it automatically from the data remains future work.

Fourth, \textsc{MultAttrEval} uses model-generated questions and answers, and its quality checks assess correctness and evidential support rather than linguistic naturalness. The method nonetheless improves over VLM prompting on human-written Qasper questions ($+0.053$ F1; Section~\ref{sec:other_datasets}) and leads GPT-4o on MCiteBench's human-origin, single-source Explanation questions ($58.12$ vs.\ $56.82$ S-F1; Appendix~\ref{app:mcite}). Its image-dense documents also contain near-duplicate or decorative images, and the single-source image regime admits one ground-truth image per item, so a method citing a visually similar alternative is scored as incorrect, and measured image precision can be depressed for over-citing methods. Such near-duplicates account for 7 of our 15 lowest-F1 items (Appendix~\ref{app:failure_modes}). Stricter image-relevance filtering and multi-image attribution are natural extensions of the benchmark.

Finally, \textsc{MultAttnAttrib} selects heads by comparing attention before and after the ground-truth evidence is replaced. We therefore interpret its attributions as intervention-informed retrieval signals: unlike purely correlational attention scores, their selection reflects sensitivity to a controlled change in the evidence. This supports evidence localization but does not constitute a complete mechanistic account of answer generation, which we leave to future work.

\bibliography{custom}

\clearpage
\appendix

\section{Artifacts and Licenses}
\textsc{MultAttrEval} is derived entirely from the publicly available
MINT-1T corpus (CC-BY-4.0), and both benchmark
construction and \textsc{MultAttnAttrib} use the open-weight
Qwen3-VL-30B-A3B-Instruct model (Apache 2.0). We additionally evaluate
InternVL3.5-30B-A3B, Gemma-4-26B-A4B, Mistral-Small-3.2-24B, and
Tomoro-ColQwen3-8B under their respective open licenses, and access
GPT-5.4 and Cohere Embed v4 through their commercial APIs in accordance
with their terms of use. All artifacts are used for research purposes
consistent with their intended use. The complete benchmark is released
under CC-BY-4.0 at \url{https://huggingface.co/datasets/dqttran/MultAttrEval},
and the full codebase for head identification, calibration, attribution,
and all evaluation scripts is released at
\url{https://github.com/dqttran/MultAttnAttrib}.

\section{MultAttrEval Dataset Statistics and Analysis}
\label{app:multattreval-data}
\begin{figure}[h!]
    \centering
    \includegraphics[width=\linewidth]{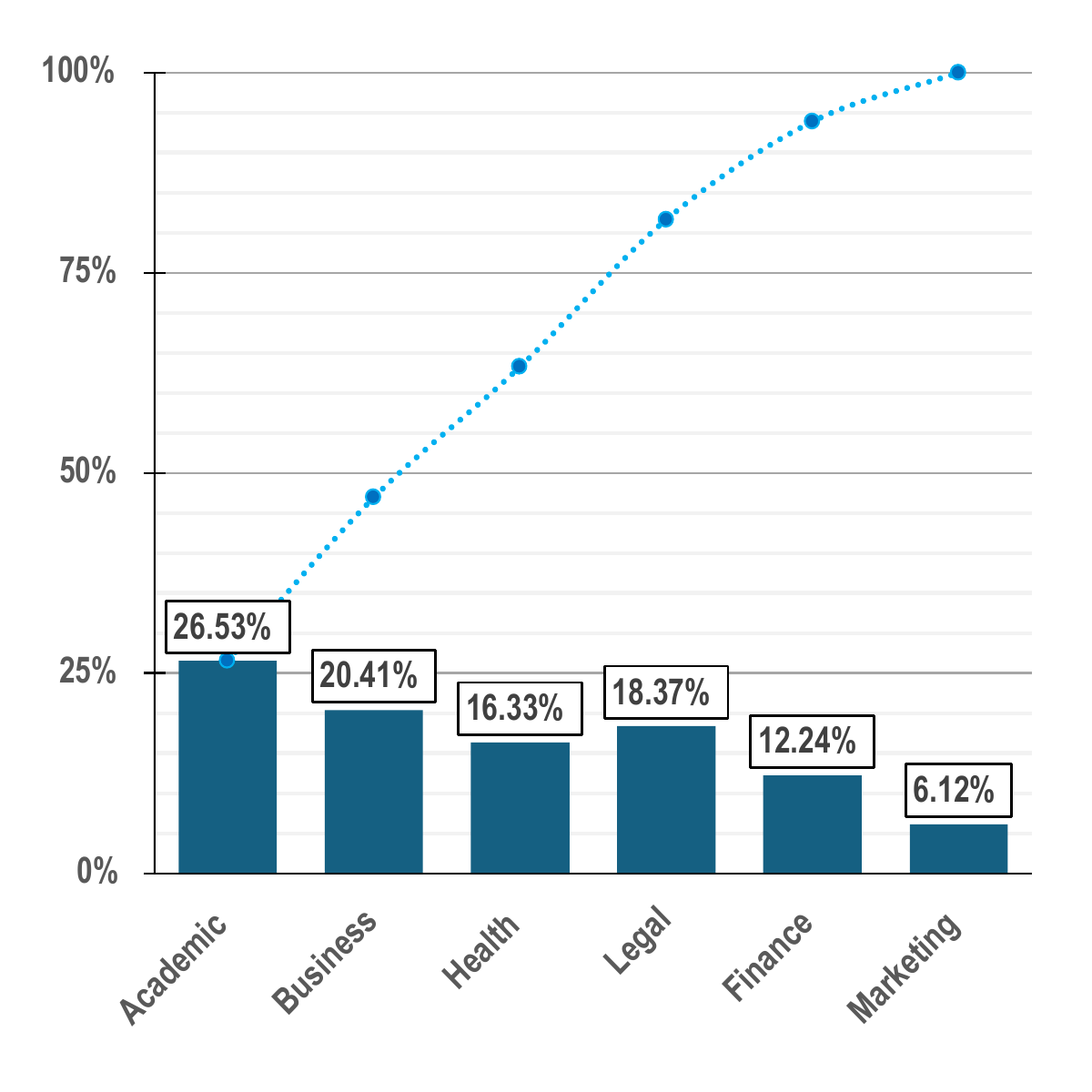}
    \caption{Distribution of MultAttrEval source documents by domain.}
    \label{fig:domain_count}
\end{figure}
\begin{table}[h!]
    \centering
    \small
    \begin{tabularx}{\linewidth}{@{}X|ccc@{}}
        \toprule
        Domain & Docs & OCR Words & Images \\
        \midrule
        Academic  & 13 & 67,667  & 162 \\
        Business  & 10 & 31,948  & 98 \\
        Finance   & 8  & 37,227  & 84 \\
        Health    & 9  & 39,753  & 95 \\
        Legal     & 6  & 19,003  & 99 \\
        Marketing & 3  & 9,794   & 33 \\
        \midrule
        Total     & 49 & 205,392 & 571 \\
        \bottomrule
    \end{tabularx}
    \caption{Table containing document statistics across domains}
\label{tab:qaa_doc_stats}
\end{table}

\begin{table}[h!]
    \centering
    \small
    \begin{tabularx}{\linewidth}{@{}X|c|cccc@{}}
        \toprule
        \multicolumn{6}{l}{\textit{\textbf{First wave}}} \\\midrule
        Domain & Docs & Text & Image & Both & Total \\
        \midrule
        Academic  & 7  & 75 & 82 & 113 & 270 \\
        Business  & 5  & 40 & 68 & 56  & 164 \\
        Health    & 5  & 46 & 54 & 86  & 186 \\
        Legal     & 2  & 12 & 41 & 7   & 60 \\
        Marketing & 1  & 3  & 10 & 5   & 18 \\
        \midrule
        Total     & 20 & 176 & 255 & 267 & 698 \\
        \midrule
        \multicolumn{6}{l}{\textit{\textbf{Second wave}}} \\\midrule
        Domain & Docs & Text & Image & Both & Total \\
        \midrule
        Academic  & 6  & 60 & 123 & 139 & 322 \\
        Business  & 5  & 50 & 145 & 67  & 262 \\
        Health    & 4  & 40 & 87  & 115 & 242 \\
        Legal     & 4  & 40 & 54  & 80  & 174 \\
        Finance   & 8  & 80 & 286 & 138 & 504 \\
        Marketing & 2  & 20 & 23  & 40  & 83 \\
        \midrule
        Total     & 29 & 290 & 718 & 579 & 1587 \\
        \bottomrule
    \end{tabularx}
    \caption{Two waves of MultAttrEval distribution across document domains and modality regimes}
    \label{tab:multattreval_dist_split}
\end{table}

\begin{table}[h!]
    \centering
    \small
    \begin{tabularx}{\linewidth}{@{}X|cc@{}}
        \toprule
        Subset & Text attr./ex. & Image attr./ex. \\
        \midrule
        Text-only &  1.00 & 0.00  \\
        Image-only &  0.00 & 1.00  \\
        Combined &  1.00 & 1.35  \\
        \bottomrule
    \end{tabularx}\\
    \begin{tabularx}{\linewidth}{@{}X|cc@{}}
    \toprule
    Subset &  Avg. Q words & Avg. A words \\
    \midrule
    Text-only &  12.9 & 16.7 \\
    Image-only &  14.1 & 13.7 \\
    Combined &  14.3 & 23.7 \\
    \bottomrule
\end{tabularx}
\caption{Analysis of QAA statistics across regimes}
\label{tab:qaa_stats}
\end{table}
\begin{figure}[h!]
    \centering
    \includegraphics[width=\linewidth]{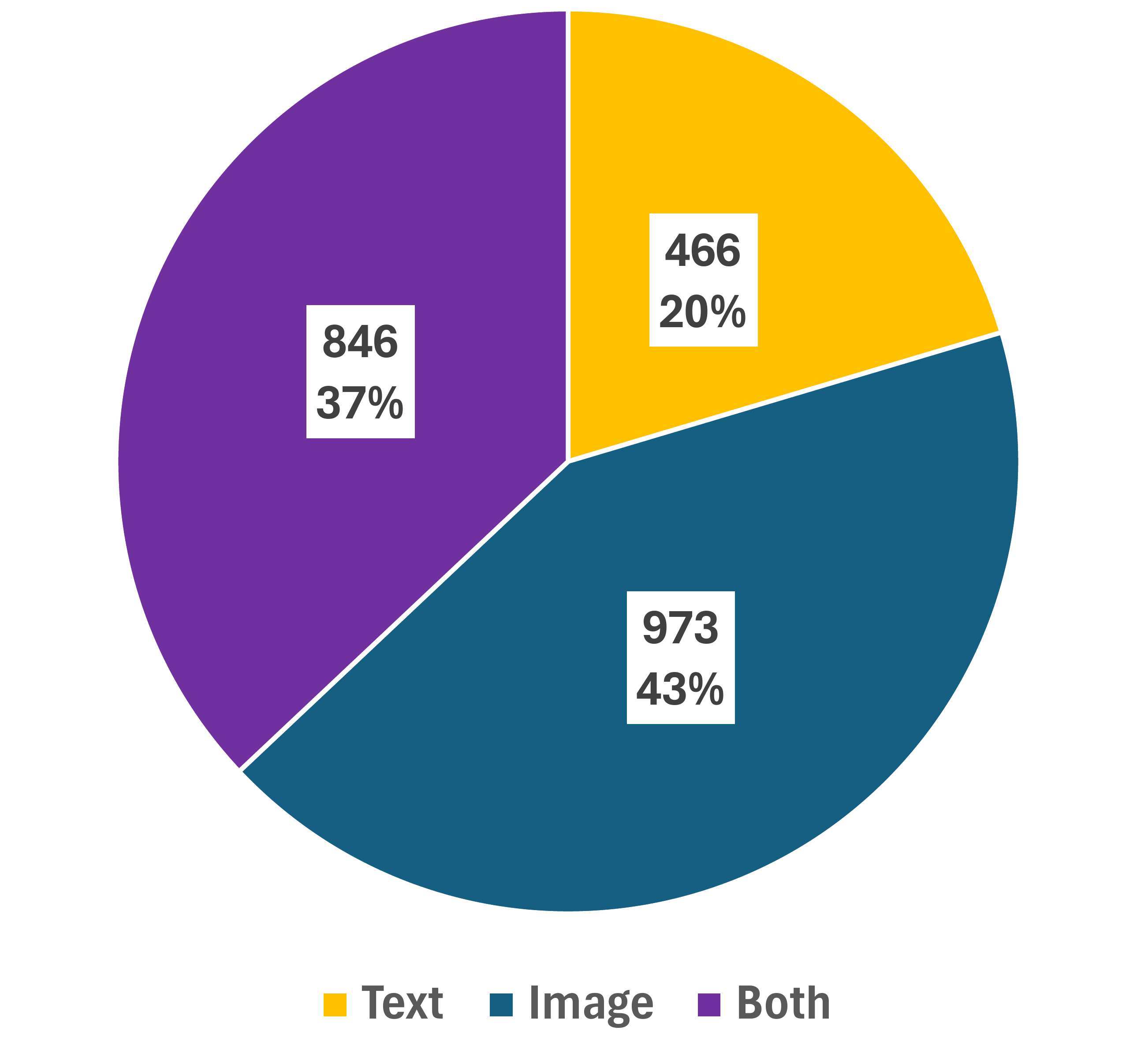}
    \caption{Distribution of \textsc{MultAttrEval} QAA items by attribution regime.}
    \label{fig:qaa_domain}
\end{figure}
\begin{figure}[h!]
    \centering
    \includegraphics[width=\linewidth]{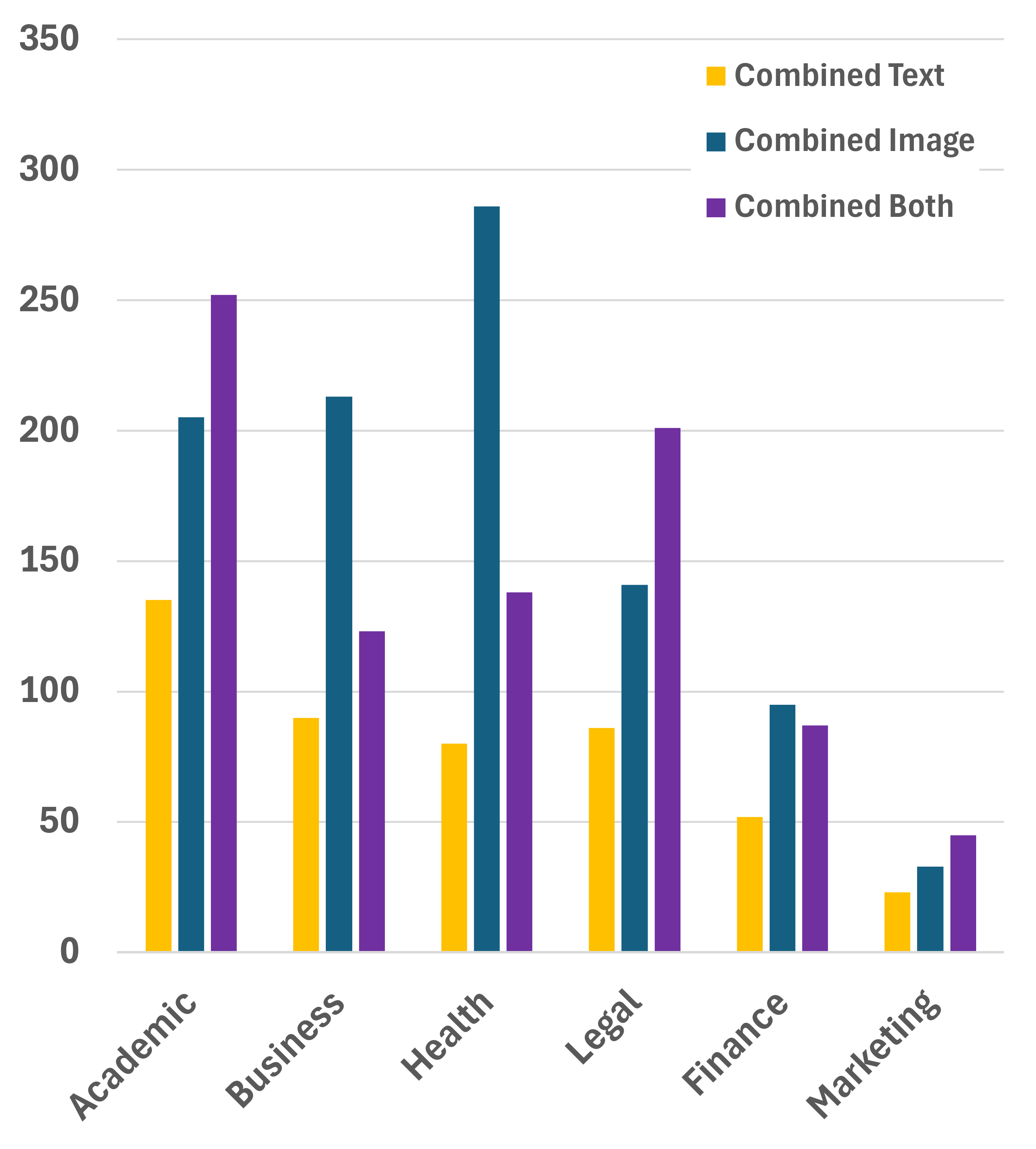}
    \caption{Distribution of \textsc{MultAttrEval} QAA items by attribution regime and document domain.}
    \label{fig:qaa_split_domain}
\end{figure}
\begin{figure}[h!]
    \centering
    \includegraphics[width=\linewidth]{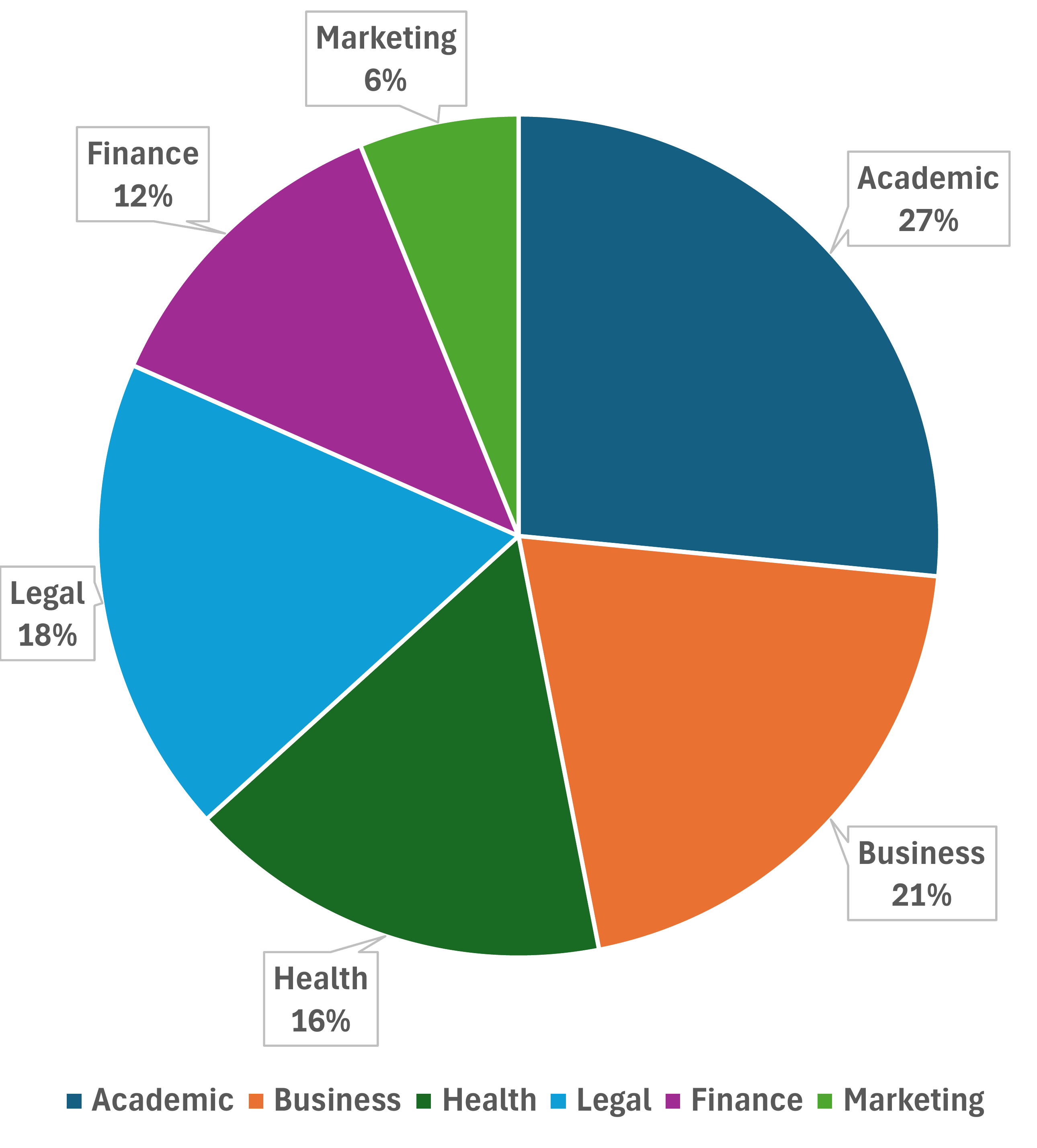}
    \caption{Pie Distribution of combined \textsc{MultAttrEval} QAA items by document domains.}
    \label{fig:qaa_combined_domain}
\end{figure}
\subsection{Annotation Protocol}
\label{app:annotation}

\paragraph{Annotators.} Six student researchers served as annotators. All are fluent English readers with familiarity in at least one of the benchmark's document domains. All volunteered without compensation. Each sampled QAA was judged independently by two of the six annotators, who were instructed not to discuss items with one another, so the agreement rates in Tables~\ref{tab:human_text}--\ref{tab:human_mult} reflect independent judgments rather than negotiated consensus. Annotation was performed in a browser-based tool that presented the document, question, answer, and ground-truth evidence (rendered text span plus any referenced images), autosaved progress locally, and exported judgments as CSV. Annotators were told to expect roughly two to three minutes per item.

\paragraph{Criteria.} Annotators answered two binary questions per item, under
the following instructions given verbatim:

\begin{lstlisting}[caption={Annotator instructions for \texttt{answer\_correct}.}]
Is the answer a correct answer to the question, given the document?

Yes: factually consistent with the document and actually answers the question.
     Minor phrasing awkwardness or extra detail is fine.
No:  contradicts the document, is unsupported by it, or does not answer the
     question asked.
\end{lstlisting}

\begin{lstlisting}[caption={Annotator instructions for \texttt{attribution\_supports}.}]
Does the shown ground-truth evidence (text span and/or images) support the
answer?

Yes: it supports at least one fact in the answer. Partial validation is
     sufficient. For multimodal items: the text supports at least one fact AND
     the image is relevant (they may support different parts).
No:  it is unrelated to the answer, directly contradicts any part of it
     (contradictions weigh heavily), or a referenced image could not plausibly
     ground anything in the answer.

Judge the evidence as given. Do not search the document for a better span.
Err on the side of support for borderline cases. If an image fails to render,
note it and leave that judgment empty.
\end{lstlisting}

\paragraph{Interpreting the human judge rates.} First, the reported rate is best read as an upper bound on the fraction of items whose ground-truth attribution is fully sufficient because \texttt{attribution\_supports} is a partial-support criterion with an explicit instruction to resolve borderline cases in favor of support. Second, annotators judged only the evidence attached to each item and were not asked whether a better span existed elsewhere in the document; the study therefore validates the attributions we ship rather than the completeness of the evidence selection.

\section{MultAttrEval Quality Verification}
\subsection{Data Quality Gates}
\label{app:multattreval_qaa}
\begin{table}[h!]
\centering
\small
\begin{tabularx}{\linewidth}{X|X}
\toprule
\textbf{Criterion} & \textbf{Description} \\
\midrule
Quality thresholding & Minimum verifier score to exclude low-value QA pairs (Listing~\ref{lst:quality}) \\\midrule
Attribution support & Filters items whose attribution isn't used to generate the answer \\\midrule
Evidence consistency & Evidence span is non-empty, 12--25 words, and an exact substring of the source paragraph \\\midrule
Cross-chunk evidence uniqueness & Rejects spans crossing more paragraph chunks than a set threshold \\
\bottomrule
\end{tabularx}
\caption{\textbf{Automated: }QAA acceptance criteria for \textbf{text-only} items.}
\label{tab:qaa-filtering-text}
\end{table}

\begin{table}[h!]
\centering
\small
\begin{tabularx}{\linewidth}{X|X}
\toprule
\textbf{Criterion} & \textbf{Description} \\
\midrule
Source referencing & QA pairs reference domain topics, not the source artifact directly \\\midrule
Question triviality & Rejects questions targeting layout artifacts (arrows, boxes, callouts) \\\midrule
Answerability & Degree to which the answer is grounded in and derivable from the source (Listing~\ref{lst:img-certify}) \\
\bottomrule
\end{tabularx}
\caption{\textbf{Automated: }QAA acceptance criteria for \textbf{image-only} and {multimodal} items.}
\label{tab:qaa-filtering-image}
\end{table}

\begin{table}[h!]
\centering
\small
\begin{tabularx}{\linewidth}{X|X}
\toprule
\textbf{Criterion} & \textbf{Description} \\
\midrule
Entity verification & 1--5 scale checking that the image visually contains the entities named in the answer (Listing~\ref{lst:entity}) \\\midrule
Crossmodal synergy & 1--4 scale requiring each modality to contribute distinct information, so neither alone suffices (Listing~\ref{lst:synergy}) \\
\bottomrule
\end{tabularx}
\caption{\textbf{Automated: }Additional QAA acceptance criteria for \textbf{multimodal} items.}
\label{tab:qaa-filtering-multimodal}
\end{table}

\subsection{Human-in-Loop}
\begin{table}[h!]
    \centering
    \small
    \begin{tabularx}{\linewidth}{@{}X|ccc}
        \toprule
        & Agree (Yes) & Disagree & Agree (No)  \\\midrule
        Correct answer & 219 & 8 & 2 \\\midrule
        Attrib supports & 212 & 16 & 1 \\\bottomrule
    \end{tabularx}
    \caption{\textbf{Human-in-loop} results for Text QAA quality verification of the dataset}
    \label{tab:human_text}
\end{table}
\begin{table}[h!]
    \centering
    \small
    \begin{tabularx}{\linewidth}{@{}X|ccc}
        \toprule
        & Agree (Yes) & Disagree & Agree (No)  \\\midrule
        Correct answer & 183 & 36 & 11 \\\midrule
        Attrib supports & 203 & 20 & 7 \\\bottomrule
    \end{tabularx}
    \caption{\textbf{Human-in-loop} results for Image QAA quality verification of the dataset}
    \label{tab:human_img}
\end{table}
\begin{table}[h!]
    \centering
    \small
    \begin{tabularx}{\linewidth}{@{}X|ccc}
        \toprule
        & Agree (Yes) & Disagree & Agree (No)  \\\midrule
        Correct answer & 208 & 22 & 0 \\\midrule
        Attrib supports & 180 & 40 & 10 \\\bottomrule
    \end{tabularx}
    \caption{\textbf{Human-in-loop} results for Text + Image QAA quality verification of the dataset}
    \label{tab:human_mult}
\end{table}

\section{Prompting Baseline Design}
\label{app:baseline}
For all prompting baselines, we experimented with an open source model, Qwen3-VL-30B-A3B-Instruct, and a closed-source frontier model, GPT-5.4. Both models can ingest long documents with interleaved text and images and have been shown to perform well across multiple VQA benchmarks. Our high-level goal was to compare our method against a diverse set of attribution-generation methods, yielding a basic VLM prompting baseline for generating attributions.

Additional baselines are organized around two central questions: whether providing full document context (raster + OCR text) gives a VLM an advantage for attribution, and whether replacing visual content with text captions (effectively reducing the task to a text-only problem) is competitive. We further test each setting with and without retrieval augmentation to isolate the contribution of context compression. This yields four prompting baselines and their subsequent variants:

\paragraph{VLM}
We provide raster image data and document text, along with a batch of QAA's from the document, and prompt a Visual-Language Model to identify where the provided answer could be sourced.

\paragraph{LLM}
We provide captions for each image and document text along with a batch of QAAs from the document, and prompt a Language Model to identify where the provided answer could be sourced.

\paragraph{RAG}
Both RAG variants operate similarly: sources are embedded into a shared space, and the top sources are retrieved against the QA embedding ($k=5$). \texttt{Cohere} retrieves the top-5 text chunks and top-5 images independently, while \texttt{ColQwen} retrieves the top-5 full PDF pages.

\paragraph{MAVIS}
MAVIS~\citep{song2026mavis} is a benchmark rather than a post-hoc attribution method, so we adapt its citation format: we number every text block and figure in the full document and prompt for sentence-final \texttt{[i]} citations, retaining its native ability to cite images. Text units are the document's own paragraph chunks.

\paragraph{TracLLM}
We run TracLLM~\citep{TracLLM} with its STC score function on Qwen3-VL-30B-A3B-Instruct, over paragraph chunks and figures in reading order. Since TracLLM is text-only, we insert each figure's vision tokens at its position so it can attribute visual evidence, leaving its search unchanged.

\paragraph{M3DocRAG}
We use M3DocRAG's ColPali-v1.2 retriever~\citep{cho2024m3docrag} to retrieve the top-5 pages within each document, which then go to the same VLM prompt as \texttt{ColQwen}. The retriever is the only difference between the two.

\section{Evaluation Metrics}
\label{app:eval_metrics}
We evaluate attribution quality using precision, recall, and $\mathrm{F1}$ score.

For each item, let $G$ and $C$ denote the ground-truth and predicted citation sets,
\[
G = \mathcal{I}^* \cup \{\mathcal{T}_{i:j}\}, \qquad
C = \hat{\mathcal{I}} \cup \{\mathcal{T}_{k:\ell}\},
\]
where $\mathcal{I}^*,\, \hat{\mathcal{I}} \subseteq \mathcal{I}$ are the ground-truth and predicted image sets, $\mathcal{T}_{i:j}$ and $\mathcal{T}_{k:\ell}$ is the ground-truth and predicted text spans.

\paragraph{Citation scoring.}

Image citations are scored by exact match:
\begin{align*}
\sigma_p(I_m,\, G) &= \mathbb{1}[I_m \in \mathcal{I}^*], \\
\sigma_r(I_m,\, C) &= \mathbb{1}[I_m \in \hat{\mathcal{I}}].
\end{align*}

Text citations are scored by fuzzy substring matching. Let $s^* = \texttt{partial\_ratio}(\mathcal{T}_{k:\ell},\, \mathcal{T}_{i:j}) \in [0,1]$.

The match score is discretized to reduce sensitivity to trivial differences:
\begin{equation}
d(s^*) = \begin{cases}
1.0, & \text{if } s^* \geq 0.9 \\
0.5, & \text{if } 0.6 \leq s^* < 0.9 \\
0.0, & \text{otherwise}
\end{cases}
\label{eq:discrete}
\end{equation}

Length ratios penalize precision for over-quoting and recall for under-quoting:
\begin{align*}
    \sigma_p(\mathcal{T}_{k:\ell},\, G) &= d(s^*)\cdot\min\!\left(1,\frac{|\mathcal{T}_{i:j}|}{|\mathcal{T}_{k:\ell}|}\right),\\
    \sigma_r(\mathcal{T}_{i:j},\, C) &= d(s^*)\cdot\min\!\left(1,\frac{|\mathcal{T}_{k:\ell}|}{|\mathcal{T}_{i:j}|}\right).
\end{align*}

\paragraph{Precision and recall.} Per-item precision and recall are the mean scores over predicted and ground-truth citations, respectively:
\begin{gather*}
\mathrm{P} = \frac{1}{|C|}\sum_{c \in C}\sigma_p(c, G), \qquad \mathrm{R} = \frac{1}{|G|}\sum_{g \in G}\sigma_r(g, C) \\
\mathrm{F1} = \frac{2PR}{P + R}
\end{gather*}
$\mathrm{P}$ and $\mathrm{R}$ are macro-averaged over the dataset, and $\mathrm{F1}$ is computed from those two averages.

\paragraph{Passage-level versus pooled evidence.} We score each predicted citation independently rather than pooling an item's evidence. On \textsc{MultAttrEval} the two views coincide: recall because each item carries one gold text span and one gold image set, scored elementwise; precision because it averages over the \emph{predicted} set, and every prompting baseline and \textsc{MultAttnAttrib} emits at most one text span per item: for \textsc{MultAttnAttrib}, a consequence of $\beta_\mathrm{txt} = 1.0$ (Section~\ref{sec:implementation}); for the prompting baselines, of the single \texttt{text\_quote} field in their output schema (Appendix~\ref{app:baseline}). They diverge only where an answer rests on multiple text spans, as on the external benchmarks (Section~\ref{sec:other_datasets}).
\paragraph{Multiple text spans.} There, both $\{\mathcal{T}_{i:j}\}$ and $\{\mathcal{T}_{k:\ell}\}$ hold more than one element, and each citation is scored against the best-matching element of the opposing set:
\begin{align*}
\sigma_p(\mathcal{T}_{k:\ell},\, G) &= d(s^*)\cdot\min\!\left(1,\tfrac{|\mathcal{T}_{i^*:j^*}|}{|\mathcal{T}_{k:\ell}|}\right),\\
\sigma_r(\mathcal{T}_{i:j},\, C) &= d(s^*)\cdot\min\!\left(1,\tfrac{|\mathcal{T}_{k^*:\ell^*}|}{|\mathcal{T}_{i:j}|}\right),
\end{align*}
where $s^*$ is evaluated at the maximizing element, $\mathcal{T}_{i^*:j^*} \in G$ for precision and $\mathcal{T}_{k^*:\ell^*} \in C$ for recall. Precision still averages over $C$ and recall over $G$, so an extra unsupported span costs precision while a second correct span gains recall, neither expressible by a single-span predictor, which bounds per-item recall well below $1$ whenever $|G| > 1$. Both expressions reduce to the single-span case when each set holds one text span, leaving all \textsc{MultAttrEval} numbers unchanged.

\section{Data Expansion Consistency}
\label{app:expansion}
The expansion wave (29 documents, 1{,}587 triplets) is released as a held-out
extension and is not used anywhere in the main results. To verify that our
findings are not specific to the first-wave split, we re-run the VLM prompting
baseline and \textsc{MultAttnAttrib} (Full Document) on it. Importantly, no
recalibration is performed: the top-$k$ heads, sliding window $W$, and
thresholds $T_\mathrm{img}$ and $T_\mathrm{txt}$ are exactly those fixed on the
90-item first-wave probe set (Section~\ref{sec:head_identification}), and no
expansion-wave item enters head selection or calibration. The expansion wave is
therefore evaluated in full ($n = 1587$, in which $290$ are text-only, $718$ are
image-only, and $579$ are combined).

\begin{table}[htbp]
\centering
\small
\setlength{\tabcolsep}{3pt}
\begin{tabularx}{\linewidth}{@{}Xl|ccc@{}}
\toprule
Regime & Method & P & R & F1 \\
\midrule
\multicolumn{5}{l}{\textit{First wave (608-item test split)}} \\
\midrule
Text   & VLM  & 0.666 & 0.382 & 0.485 \\
            & Ours & 0.572 & 0.621 & \textbf{0.596} \\
Image  & VLM  & 0.477 & 0.871 & 0.617 \\
            & Ours & 0.750 & 0.804 & \textbf{0.776} \\
Text+Image  & VLM  & 0.524 & 0.465 & 0.493 \\
            & Ours & 0.626 & 0.544 & \textbf{0.582} \\
\midrule
\multicolumn{5}{l}{\textit{Expansion wave (1{,}587 items, no recalibration)}} \\
\midrule
Text   & VLM  & 0.635 & 0.357 & 0.457 \\
            & Ours & 0.528 & 0.688 & \textbf{0.597} \\
Image & VLM  & 0.478 & 0.929 & 0.631 \\
            & Ours & 0.921 & 0.964 & \textbf{0.942} \\
Text+Image  & VLM  & 0.582 & 0.493 & 0.534 \\
            & Ours & 0.715 & 0.612 & \textbf{0.660} \\
\bottomrule
\end{tabularx}
\caption{\textbf{Scale consistency check.} VLM prompting versus
\textsc{MultAttnAttrib} (Full Document) on both waves, using heads and
thresholds calibrated only on the first-wave probe set. The ordering and the
direction of the modality asymmetry are preserved at $2.6\times$ the item count
without recalibration.}
\label{tab:expansion_consistency}
\end{table}

Every margin holds and widens: image-only $+0.311$ F1 against $+0.159$ on the first wave, text-only $+0.140$ against $+0.111$, and combined $+0.126$ against $+0.089$. Text-only F1 is nearly invariant ($0.596 \rightarrow 0.597$), indicating that $T_\mathrm{txt}$ transfers without adjustment.

Absolute image-only scores rise across waves ($0.776 \rightarrow 0.942$), which
we attribute to composition rather than method behavior. Per-regime scores are
computed within regime, so what matters is the domain mix inside the image-only
pool rather than its share of the benchmark: Finance, absent from the first
wave, supplies 286 of the 718 expansion image-only items (39.8\%), while
Academic, the hardest domain in our difficulty analysis
(Appendix~\ref{app:domain_difficulty}), falls from 32.2\% to 17.1\% of
image-only items. Finance and business documents are chart- and table-heavy and
carry fewer of the near-duplicate images that penalize our single-source image
regime (Section~\ref{sec:limitations}). The VLM baseline moves in the same
direction on image-only recall ($0.871 \rightarrow 0.929$), consistent with an
easier image pool rather than a change in either method. We read
Table~\ref{tab:expansion_consistency} as evidence of stability under transfer,
not as an improved headline result.

\section{MCiteBench: Native-Protocol Results}
\label{app:mcite}
Section~\ref{sec:other_datasets} evaluates MCiteBench \citep{hu2025mcitebench} under a post-hoc protocol matched to our own (Table~\ref{tab:posthoc_external}). Here we additionally evaluate under MCiteBench's own protocol and metrics, since it generates the answer and its citations jointly rather than post-hoc. MCiteBench scores citation sets with Source-F1 (S-F1), the harmonic mean of citation-set precision and recall against the gold citation set, and Source Exact Match (S-EM), which is $1$ if the predicted citation set exactly equals the gold set and $0$ otherwise, over three categories: Explanation (single-source), Explanation (multi-source), and Locating (single-source); there is no multi-source Locating category in the benchmark. Because \textsc{MultAttnAttrib} is post-hoc, we first generate an answer with Qwen3VL-30B-A3B-Instruct and then attribute it, rather than using MCiteBench's own reference answers, which leak their own evidence identifiers. As in Section~\ref{sec:head_identification}, heads and thresholds are re-derived on a disjoint probe set drawn from MCiteBench itself, 120 items split evenly across its four source types (30 text, 30 figure, 30 table, 30 mixed).

\begin{table}[h!]
\centering
\small
\setlength{\tabcolsep}{4pt}
\begin{tabularx}{\linewidth}{@{}Xl|cc@{}}
\toprule
Category & Method & S-F1 & S-EM \\
\midrule
Explanation & \textsc{MultAttnAttrib} & \textbf{58.12} & \textbf{35.83} \\
(single-source) & Open-source ceiling & 45.13 & 35.36 \\
 & GPT-4o & 56.82 & 24.50 \\
\midrule
Explanation & \textsc{MultAttnAttrib} & 57.18 & 18.34 \\
(multi-source) & Open-source ceiling & 57.65 & 16.86 \\
 & GPT-4o & \textbf{67.56} & \textbf{21.27} \\
\midrule
Locating & \textsc{MultAttnAttrib} & \textbf{91.34} & \textbf{82.09} \\
(single-source) & Open-source ceiling & 68.86 & 61.48 \\
 & GPT-4o & 85.74 & 69.45 \\
\bottomrule
\end{tabularx}
\caption{MCiteBench under its native protocol. Open-source ceiling is the best score per column across InternVL2.5-8B/78B and Qwen2-VL-7B/72B.}
\label{tab:mcite_native}
\end{table}

\textsc{MultAttnAttrib} leads both GPT-4o and the open-source ceiling on Explanation (single-source) and on Locating, improving over GPT-4o by $5.6$ S-F1 and $12.6$ S-EM on the latter. It trails GPT-4o on Explanation (multi-source), where roughly half the items require citing multiple text or image sources -- a setting \textsc{MultAttrEval} does not cover and our calibration does not target, since we calibrate for one supporting span per item (Section~\ref{sec:limitations}).

\section{MultAttnAttrib}
\label{app:algorithms}

In every pass, attention to the first (sink) token is zeroed. In Algorithms~\ref{alg:head_meanattn} and~\ref{alg:head_cma}, the ground-truth positions $G_i$ are restricted to content tokens, excluding stopwords and punctuation.

{\paragraph{Probe size and transfer.} The 30- and 60-item probes (Section~\ref{sec:results}) are stratified draws from the 90-item probe (10 and 20 items per family); for each, heads and thresholds are re-derived from scratch with $k = 8$, and the model is evaluated on the same 608-item test split. For transfer, heads and thresholds are derived on the MCiteBench or Qasper probe and applied to \textsc{MultAttrEval} unchanged; all other settings follow Section~\ref{sec:implementation}. The Qasper-derived heads are evaluated on text-only items, since a text-only probe yields no image threshold. Both are compared with VLM prompting on the same backbone (Table~\ref{tab:main_results}).}

\begin{algorithm}[h!]
\caption{\textsc{MultAttnAttrib}: Head Identification (MAS)}
\label{alg:head_meanattn}
\begin{algorithmic}[1]
\small
\Require $g_\phi$ (language model), $\{(\mathbf{x}_i, G_i, Q_i, D_i)\}_{i=1}^N$ (probe set), $k$ (number of heads)
\State $S \leftarrow \mathbf{0}^{L \times H}$
\For{$i \leftarrow 1, \ldots, N$}
    \State $A \leftarrow g_\phi(\mathbf{x}_i, Q_i)$
    \For{$(l,h)$}
        \State $\bar{a}(X) \leftarrow \operatorname{mean}_{q\in Q_i} A_{l,h,q}(X)$
        \State $r \leftarrow \bar{a}(G_i) \,/\, \big(\bar{a}(G_i) + \bar{a}(D_i{\setminus}G_i)\big)$
        \State $w \leftarrow \max(0,\; 1 - H(A_{l,h,\cdot}|_{D_i})\,/\,\log|D_i|)$
        \State $S[l,h] \mathrel{+}= r \cdot w$
    \EndFor
\EndFor
\State $\mathcal{H} \leftarrow \text{arg\,max}_{l,h}^k \, S[l,h]$
\State \Return $\mathcal{H}$
\Statex
\end{algorithmic}
\end{algorithm}

\begin{algorithm}[h!]
\caption{\textsc{MultAttnAttrib}: Head Identification (CMA)}
\label{alg:head_cma}
\begin{algorithmic}[1]
\small
\Require $g_\phi$ (language model), $\{(\mathbf{x}_i, G_i, Q_i, D_i)\}_{i=1}^N$ (probe set), $k$ (number of heads)
\State $S \leftarrow \mathbf{0}^{L \times H}$
\For{$i \leftarrow 1, \ldots, N$}
    \State $\tilde{\mathbf{x}}_i \leftarrow \textsc{Corrupt}(\mathbf{x}_i)$
    \State $A^{\mathrm{clean}} \leftarrow g_\phi(\mathbf{x}_i, Q_i)$;\quad
           $A^{\mathrm{corrupt}} \leftarrow g_\phi(\tilde{\mathbf{x}}_i, Q_i)$
    \For{$(l,h)$}
        \State $\mathrm{IE} \leftarrow \operatorname{mean}_{q\in Q_i}\bigl[A^{\mathrm{clean}}_{l,h,q}(G_i) - A^{\mathrm{corrupt}}_{l,h,q}(G_i)\bigr]$
        \State $w \leftarrow \max(0,\; 1 - H(A^{\mathrm{clean}}_{l,h,\cdot}|_{D_i})\,/\,\log|D_i|)$
        \State $S[l,h] \mathrel{+}= \mathrm{IE} \cdot w$
    \EndFor
\EndFor
\State $\mathcal{H} \leftarrow \text{arg\,max}_{l,h}^k \, S[l,h]$
\State \Return $\mathcal{H}$
\Statex
\end{algorithmic}
\end{algorithm}

\begin{algorithm}[h!]
\caption{\textsc{MultAttnAttrib}: Calibration}
\label{alg:calibration}
\begin{algorithmic}[1]
\small
\Require $\{v^{\text{img}}_i, v^{\text{txt}}_i\}$ (probe attribution scores), $\{\mathcal{G}_i\}$ (ground-truth modality labels)
\For{$m \in \{\text{img},\,\text{txt}\}$}
    \State $(V^+_m,\, V^-_m) \leftarrow \textsc{Split}(\{v^m_i\},\, \{\mathcal{G}_i\})$
    \State $T_m \leftarrow \arg\max_{T}\; \mathrm{F1}(V^+_m,\, V^-_m,\, T)$
\EndFor
\State \Return $T_\text{img}\,,\, T_\text{txt}$
\end{algorithmic}
\end{algorithm}

\section{Comparing \textsc{GPT-5.4} to \textsc{Qwen3-VL-30B}}
\label{app:qwen_gpt}

\begin{table}[h!]
\centering
\small
\begin{tabularx}{\linewidth}{@{}X|ccc@{}}
    \toprule
    Method (Text) & Precision & Recall & F1 \\
    \midrule
    \multicolumn{4}{l}{\textit{GPT-5.4}} \\
    \midrule
    VLM & 0.666 & 0.539 & 0.596 \\
    LLM & 0.722 & 0.614 & 0.664 \\
    Cohere + VLM & 0.915 & 0.720 & 0.806 \\
    ColQwen + VLM & 0.701 & 0.553 & 0.618 \\
    Cohere + LLM & \textbf{0.928} & \textbf{0.851} & \textbf{0.888} \\
    \midrule
    \multicolumn{4}{l}{\textit{Qwen3-VL-30B-A3B-Instruct}} \\
    \midrule
    VLM & 0.666 & 0.382 & 0.485 \\
    LLM & 0.696 & 0.402 & 0.510 \\
    Cohere + VLM & 0.883 & 0.380 & 0.531 \\
    ColQwen + VLM & 0.721 & 0.376 & 0.494 \\
    Cohere + LLM & 0.877 & 0.415 & 0.563 \\
    \bottomrule
\end{tabularx}
\caption{Text regime metrics for GPT and Qwen3-VL}
\label{tab:qwen_gpt_txt}
\end{table}
\begin{table}[h!]
\centering
\small
\begin{tabularx}{\linewidth}{@{}X|ccc@{}}
    \toprule
    Method (Image) & Precision & Recall & F1 \\
    \midrule
    \multicolumn{4}{l}{\textit{GPT-5.4}} \\
    \midrule
    VLM & 0.653 & 0.831 & 0.732 \\
    LLM & 0.689 & 0.831 & 0.754 \\
    Cohere + VLM & 0.698 & 0.840 & \textbf{0.763} \\
    ColQwen + VLM & 0.593 & 0.667 & 0.628 \\
    Cohere + LLM & \textbf{0.702} & 0.827 & 0.759 \\
    \midrule
    \multicolumn{4}{l}{\textit{Qwen3-VL-30B-A3B-Instruct}} \\
    \midrule
    VLM & 0.477 & \textbf{0.871} & 0.617 \\
    LLM & 0.403 & 0.813 & 0.539 \\
    Cohere + VLM & 0.427 & 0.813 & 0.560 \\
    ColQwen + VLM & 0.400 & 0.698 & 0.508 \\
    Cohere + LLM & 0.466 & 0.858 & 0.604 \\
    \bottomrule
\end{tabularx}
\caption{Image regime metrics for GPT and Qwen3-VL}
\label{tab:qwen_gpt_img}
\end{table}
\begin{table}[h!]
\centering
\small
\begin{tabularx}{\linewidth}{@{}X|ccc@{}}
    \toprule
    Method (Text + Image) & Precision & Recall & F1 \\
    \midrule
    \multicolumn{4}{l}{\textit{GPT-5.4}} \\
    \midrule
    VLM & 0.583 & 0.454 & 0.511 \\
    LLM & 0.713 & 0.446 & 0.549 \\
    Cohere + VLM & 0.856 & 0.502 & 0.633 \\
    ColQwen + VLM & 0.627 & 0.433 & 0.512 \\
    Cohere + LLM & \textbf{0.918} & \textbf{0.519} & \textbf{0.663} \\
    \midrule
    \multicolumn{4}{l}{\textit{Qwen3-VL-30B-A3B-Instruct}} \\
    \midrule
    VLM & 0.524 & 0.465 & 0.493 \\
    LLM & 0.552 & 0.396 & 0.461 \\
    Cohere + VLM & 0.713 & 0.453 & 0.554 \\
    ColQwen + VLM & 0.581 & 0.484 & 0.528 \\
    Cohere + LLM & 0.725 & 0.459 & 0.562 \\
    \bottomrule
\end{tabularx}
\caption{Multimodal regime metrics for GPT and Qwen3-VL}
\label{tab:qwen_gpt_mult}
\end{table}

\section{Comparing GPT-5.4 to \textsc{MultAttnAttrib}}
\label{app:gpt_comparison}
\begin{table}[h!]
\centering
\small
\begin{tabularx}{\linewidth}{@{}X|ccc@{}}
\toprule
Method (Text) & Precision & Recall & F1 \\
\midrule
VLM & 0.666 & 0.539 & 0.596 \\
LLM & 0.722 & 0.614 & 0.664 \\
Cohere + VLM & 0.915 & 0.720 & 0.806 \\
ColQwen + VLM & 0.701 & 0.553 & 0.618 \\
Cohere + LLM & \textbf{0.928} & \textbf{0.851} & \textbf{0.888} \\
\midrule
Cohere + \textsc{MultAttnAttrib} & 0.614 & 0.726 & 0.665 \\
\midrule
\multicolumn{4}{l}{\textit{\% Change from GPT to \textsc{MultAttnAttrib}}} \\
\midrule
$\Delta$ VLM & \color{red}$-$7.8\% & \color{darkpastelgreen}+34.7\% & \color{darkpastelgreen}+11.6\% \\
$\Delta$ LLM & \color{red}$-$15.0\% & \color{darkpastelgreen}+18.2\% & \color{darkpastelgreen}+0.2\% \\
$\Delta$ Cohere + VLM & \color{red}$-$32.9\% & \color{darkpastelgreen}+0.8\% & \color{red}$-$17.5\% \\
$\Delta$ ColQwen + VLM & \color{red}$-$12.4\% & \color{darkpastelgreen}+31.3\% & \color{darkpastelgreen}+7.6\% \\
$\Delta$ Cohere + LLM & \color{red}$-$33.8\% & \color{red}$-$14.7\% & \color{red}$-$25.1\% \\
\bottomrule
\end{tabularx}
\caption{Text regime metrics for GPT and \textsc{MultAttnAttrib}}
\label{tab:gpt_comp_txt}
\end{table}

\begin{table}[h!]
\centering
\small
\begin{tabularx}{\linewidth}{@{}X|ccc@{}}
\toprule
Method (Image) & Precision & Recall & F1 \\
\midrule
VLM & 0.653 & 0.831 & 0.732 \\
LLM & 0.689 & 0.831 & 0.754 \\
Cohere + VLM & 0.698 & \textbf{0.840} & 0.763 \\
ColQwen + VLM & 0.593 & 0.667 & 0.628 \\
Cohere + LLM & 0.702 & 0.827 & 0.759 \\
\midrule
Cohere + \textsc{MultAttnAttrib} & \textbf{0.749} & 0.827 & \textbf{0.786} \\
\midrule
\multicolumn{4}{l}{\textit{\% Change from GPT to \textsc{MultAttnAttrib}}} \\
\midrule
$\Delta$ VLM & \color{darkpastelgreen}+14.7\% & \color{red}$-$0.5\% & \color{darkpastelgreen}+7.4\% \\
$\Delta$ LLM & \color{darkpastelgreen}+8.7\% & \color{red}$-$0.5\% & \color{darkpastelgreen}+4.2\% \\
$\Delta$ Cohere + VLM & \color{darkpastelgreen}+7.3\% & \color{red}$-$1.5\% & \color{darkpastelgreen}+3.0\% \\
$\Delta$ ColQwen + VLM & \color{darkpastelgreen}+26.3\% & \color{darkpastelgreen}+24.0\% & \color{darkpastelgreen}+25.2\% \\
$\Delta$ Cohere + LLM & \color{darkpastelgreen}+6.7\% & \color{darkpastelgreen}+0.0\% & \color{darkpastelgreen}+3.6\% \\
\bottomrule
\end{tabularx}
\caption{Image regime metrics for GPT and \textsc{MultAttnAttrib}}
\label{tab:gpt_comp_img}
\end{table}

\begin{table}[h!]
\centering
\small
\begin{tabularx}{\linewidth}{@{}X|ccc@{}}
\toprule
Method (Text + Image) & Precision & Recall & F1 \\
\midrule
VLM & 0.583 & 0.454 & 0.511 \\
LLM & 0.713 & 0.446 & 0.549 \\
Cohere + VLM & 0.856 & 0.502 & 0.633 \\
ColQwen + VLM & 0.627 & 0.433 & 0.512 \\
Cohere + LLM & \textbf{0.918} & 0.519 & \textbf{0.663} \\
\midrule
Cohere + \textsc{MultAttnAttrib} & 0.643 & \textbf{0.564} & 0.601 \\
\midrule
\multicolumn{4}{l}{\textit{\% Change from GPT to \textsc{MultAttnAttrib}}} \\
\midrule
$\Delta$ VLM & \color{darkpastelgreen}+10.3\% & \color{darkpastelgreen}+24.2\% & \color{darkpastelgreen}+17.6\% \\
$\Delta$ LLM & \color{red}$-$9.8\% & \color{darkpastelgreen}+26.5\% & \color{darkpastelgreen}+9.5\% \\
$\Delta$ Cohere + VLM & \color{red}$-$24.9\% & \color{darkpastelgreen}+12.4\% & \color{red}$-$5.1\% \\
$\Delta$ ColQwen + VLM & \color{darkpastelgreen}+2.6\% & \color{darkpastelgreen}+30.3\% & \color{darkpastelgreen}+17.4\% \\
$\Delta$ Cohere + LLM & \color{red}$-$30.0\% & \color{darkpastelgreen}+8.7\% & \color{red}$-$9.4\% \\
\bottomrule
\end{tabularx}
\caption{Multimodal regime metrics for GPT and \textsc{MultAttnAttrib}}
\label{tab:gpt_comp_mult}
\end{table}

\section{Domain Difficulty Analysis}
\label{app:domain_difficulty}
\begin{figure}[h!]
        \centering
        \includegraphics[width=\linewidth]{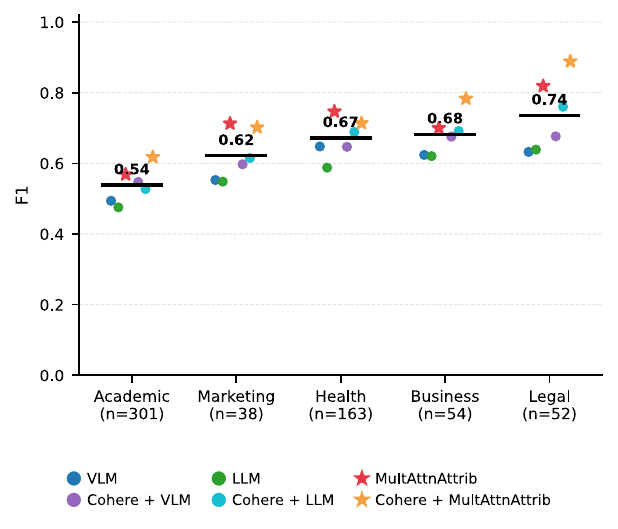}
        \caption{\textbf{Domain Difficulty Chart} F1 scores for each of our regimes, grouped by document domain and method used, ordered from hardest to easiest domains (all modalities pooled)}
        \label{fig:f1_domain}
\end{figure}

\paragraph{Domains have no tangible performance impact on intra-baseline relationships.} Generally, the VLM and LLM baselines perform the worst, with Cohere + VLM and Cohere + LLM being similarly better, and \textsc{MultAttnAttrib}, along with its Cohere variant, generally outperforming other methods. Changing the document type does not affect the relationships among the baselines, indicating that our baseline implementations are robust and impartial across the domains tested in our experiments.

\paragraph{Academic documents are consistently more difficult to generate attributions for.} Academic documents have a regime-wide unified F1 score of 0.54, with marketing, the second worst-performing domain, seeing an approximate 8 points increase in F1 scores. This is the highest domain-to-domain jump for unified F1 scores. This disparity is likely due to the structure of academic/research documents. Redundancy in academic documents is common, as reference material is sparsely distributed, cross-referential, and frequently restated or paraphrased. As a result, we often see over-attribution in a QA pairing, leading to inaccuracies when comparing against the more lightweight ground-truth attributions and to poor F1 results, as seen in Figure~\ref{fig:f1_domain}.

\paragraph{Legal documents tend to receive more accurate attributions than other domains.} We see that there is a 6 points domain-to-domain jump in F1 metrics between Legal (with an F1 score of 0.74) and Business. This suggests that the legal domain is relatively easier to generate accurate attributions for. This is because references are densely structured within specific clauses, claims, laws, or cases. This allows for fine-grained attributions for QA pairings (attributions that our baseline can locate with more ease) and creates opportunities for better fine-grained attributions.

\section{Baseline Findings}
\paragraph{Switching to image captions improves performance in GPT baselines, with mixed results in the Qwen baselines.} In the GPT case (Appendix~\ref{app:qwen_gpt}), we observed improvements particularly in precision for the text and multimodal regimes and somewhat in the image-only regime. In the Qwen case (Table \ref{tab:main_results} and Appendix \ref{app:qwen_gpt}), we see a slight boost in text-only regimes, but degradation in image-only and text + image QAAs. This asymmetry suggests that the open-source model leans more heavily on fine-grained visual representations and is less able to perform attribution reasoning over abstracted textual descriptions of images. This observation directly motivates MultAttnAttrib's design: rather than mediating images through captions, we read attribution signals off attention over image patches, where the fine-grained visual evidence is already encoded.

\paragraph{RAG generally improves metrics in comparison to direct inference, with gains depending on the regime and model used.} For GPT-5.4, layering Cohere-based RAG on top of the LLM nearly closes the gap on text-only attribution, but for Qwen, the same intervention yields only modest gains, even harming performance in the image-only regime. We hypothesize this asymmetry arises because retrieval preselects evidence into a smaller candidate pool, which a stronger generator can exploit but a weaker one cannot. Replacing Cohere with ColQwen as the retriever further degrades performance across all methods and splits, indicating that retrieval \emph{quality}, not just its addition, drives the gains we observed.

\paragraph{Multimodal attribution is challenging and resists frontier gains.} Taking the strongest baseline within each split, GPT-5.4 (Appendix \ref{app:qwen_gpt}) outperforms Qwen3-VL-30B by 32.5 F1 points on text-only and 14.6 points on image-only, but only 10.1 points on combined attribution. The pattern persists under direct baseline comparisons: with VLM, the GPT--Qwen F1 gap is 11.1 points on text-only and 11.5 on image-only, but collapses to just 1.8 on combined attribution (Table \ref{tab:main_results}). This suggests that combined attribution exposes a difficulty distinct from those addressed by scale alone, arbitrating between modalities and aggregating partial evidence from each, which current frontier-generation methods do not resolve on their own.

\section{LLM Judge Analysis}
\label{app:llm_judge}
\subsection{Judge Setup} To complement token-overlap metrics, we additionally evaluate attribution quality using a multi-judge LLM panel. Each (question, answer, answer\_part, citation) tuple is scored by a panel of three GPT-4o judges, each assigned a distinct deliberation persona: a balanced evaluator, a detail-focused critic, and a consensus mediator. Judges share a discussion history and deliberate for up to two rounds, with early termination upon unanimous consensus; the final decision is determined by majority vote. A citation is judged as \textit{supportive} if it grounds at least one fact in the answer component, and as \textit{non-supportive} if it contradicts or is entirely unrelated to the attributed claim. We report \textbf{Relevance} and \textbf{Support} as the proportions of citations judged supportive for each method across attribution regimes.

\subsection{Judge Results}
\begin{table}[h!]
\centering
\small
\begin{tabularx}{\linewidth}{@{\extracolsep{\fill}}Xcc@{}}
\toprule
Method (Text) & Relevance & Support \\
\midrule
\textit{Qwen3-VL-30B-A3B-Instruct} & &\\
\midrule
VLM & 0.712 & 0.623 \\
LLM & \textbf{0.741} & 0.641 \\
Cohere + VLM & 0.722 & 0.634 \\
Cohere + LLM & 0.724 & 0.651 \\
\midrule
\textsc{MultAttnAttrib} (Full)& 0.691 & \textbf{0.741} \\
\bottomrule
\end{tabularx}
\caption{Text regime LLM Judge results for Qwen and \textsc{MultAttnAttrib}}
\end{table}
\begin{table}[h!]
\centering
\small
\begin{tabularx}{\linewidth}{@{\extracolsep{\fill}}Xcc@{}}
\toprule
Method (Image)& Relevance & Support \\
\midrule
\textit{Qwen3-VL-30B-A3B-Instruct} & &\\
\midrule
VLM & 0.511 & 0.794 \\
LLM & 0.498 & 0.776 \\
Cohere + VLM & 0.495 & 0.781 \\
Cohere + LLM & 0.486 & 0.793 \\
\midrule
\textsc{MultAttnAttrib} (Full) & \textbf{0.561} & \textbf{0.831} \\
\bottomrule
\end{tabularx}
\caption{Image regime LLM Judge results for Qwen and \textsc{MultAttnAttrib}}
\end{table}
\begin{table}[h!]
\centering
\small
\begin{tabularx}{\linewidth}{@{\extracolsep{\fill}}Xcc@{}}
\toprule
Method (Text + Image) & Relevance & Support \\
\midrule
\textit{Qwen3-VL-30B-A3B-Instruct} & & \\
\midrule
VLM & 0.534 & 0.498 \\
LLM & 0.598 & 0.486 \\
Cohere + VLM & 0.546 & 0.489 \\
Cohere + LLM & 0.574 & 0.511 \\
\midrule
\textsc{MultAttnAttrib} (Full) & \textbf{0.598} & \textbf{0.523} \\
\bottomrule
\end{tabularx}
\caption{Multimodal regime LLM Judge results for Qwen and \textsc{MultAttnAttrib}}
\end{table}

\section{Head Analysis}
\label{app:head_analysis}
\paragraph{Setup.} We score each of the $N = L \times H = 48 \times 32 = 1536$ attention heads under two methods: Mean Attention Scoring and CMA Scoring (discussed in Section~\ref{sec:head_identification}). Both produce score matrices $\mathbf{S}^{\text{img}}, \mathbf{S}^{\text{txt}} \in \mathbb{R}^{L \times H}$, which we use to measure cross-modal agreement via IoU and Spearman's rank correlation (Figure~\ref{fig:head_agreement}).

\paragraph{Metrics.}
Let $\mathcal{H}^{\text{img}}_k$ and $\mathcal{H}^{\text{txt}}_k$ denote the top-$k$ image and text head sets, respectively.
\begin{equation*}
    \text{IoU}(k) \;=\;
    \frac{|\mathcal{H}^{\text{img}}_k \cap \mathcal{H}^{\text{txt}}_k|}
         {|\mathcal{H}^{\text{img}}_k \cup \mathcal{H}^{\text{txt}}_k|}.
\end{equation*}
Let $\mathcal{U}_k = \mathcal{H}^{\text{img}}_k \cup \mathcal{H}^{\text{txt}}_k$ and $r_i^{\text{img}}, r_i^{\text{txt}}$ be the rank of the head $i$'s score within $\mathcal{U}_k$ under specified modality. Spearman's rank correlation over this union measures how similar text and image modalities order these heads:
\begin{equation*}
    \rho_k \;=\; 1 - \frac{6\displaystyle\sum_{i \in \mathcal{U}_k}
    \!\bigl(r_i^{\text{img}} - r_i^{\text{txt}}\bigr)^2}
    {|\mathcal{U}_k|\bigl(|\mathcal{U}_k|^2 - 1\bigr)}.
\end{equation*}\\

\label{app:score_heatmap}
\begin{figure}[h!]
    \centering
    \includegraphics[width = \linewidth]{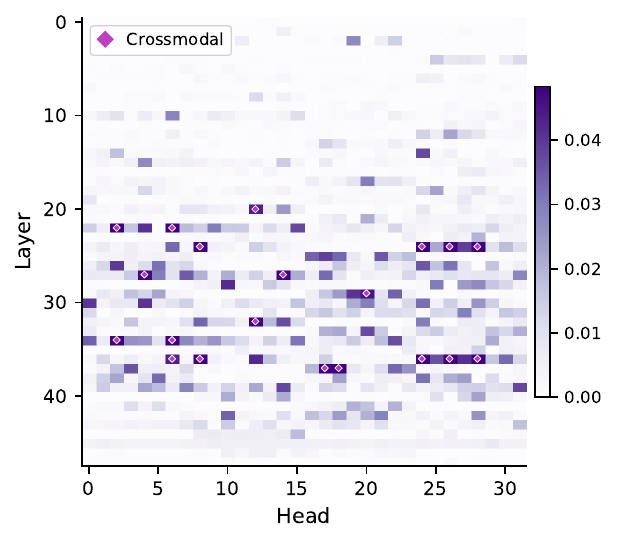}
    \caption{\textbf{Score heatmaps: }CMA attribution score heatmap for heads that jointly attend to both image and text. Diamonds mark the top-20 cross-modal heads.}
    \label{fig:score_heatmap_crm}
\end{figure}
\begin{figure}[h!]
    \centering
    \includegraphics[width=\linewidth]{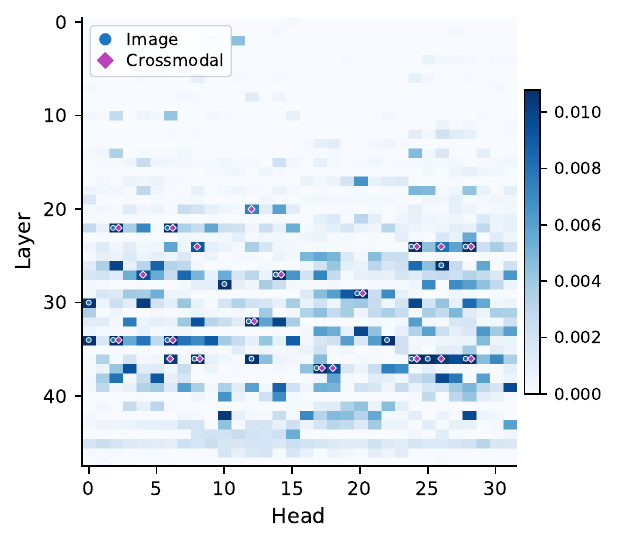}
    \caption{\textbf{Score heatmaps: }CMA attribution score heatmap for heads that attend to image sources. Circles mark the top-20 image heads, and diamonds mark the top-20 cross-modal heads.}
    \label{fig:score_heatmap_img}
\end{figure}
\begin{figure}[h!]
    \centering
    \includegraphics[width=\linewidth]{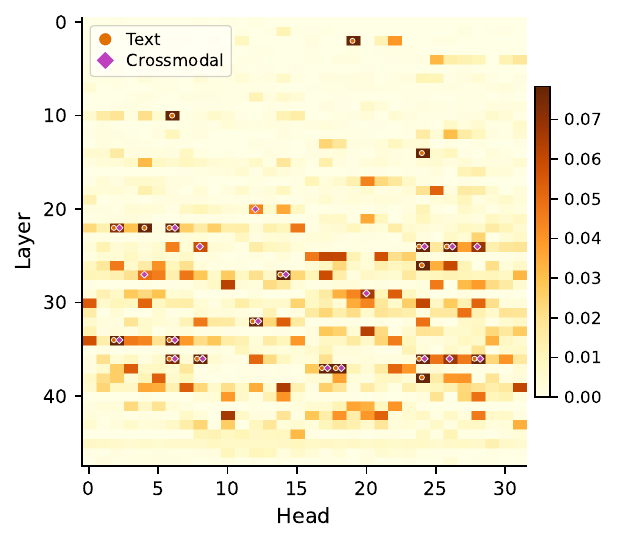}
    \caption{\textbf{Score heatmaps: }CMA attribution score heatmap for heads that attend to text sources. Circles mark the top-20 text heads, and diamonds mark the top-20 cross-modal heads.}
    \label{fig:score_heatmap_txt}
\end{figure}
\begin{figure}[h!]
    \centering
    \includegraphics[width=\linewidth]{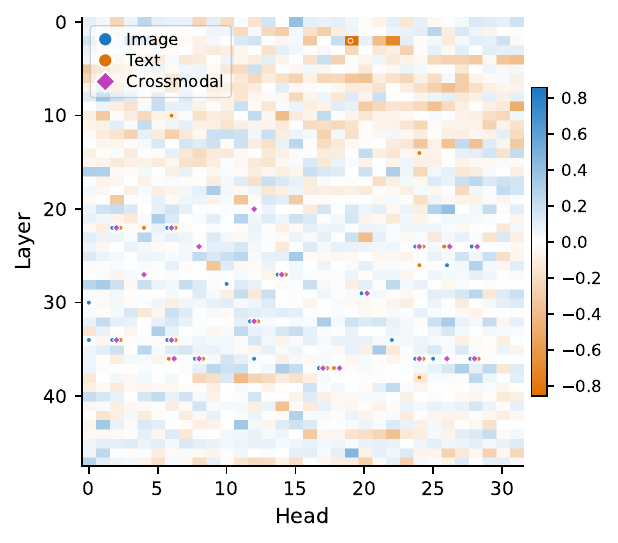}
    \caption{\textbf{Score heatmaps: }Relative modality specialization of CMA-scored heads, measured as the normalized rank difference $(r_{\text{img}} - r_{\text{txt}}) / N$. Blue cells indicate image-dominant heads, while orange cells indicate text-dominant heads.}
    \label{fig:score_heatmap_diff}
\end{figure}

\begin{figure}[h!]
    \centering
    \includegraphics[width=\linewidth]{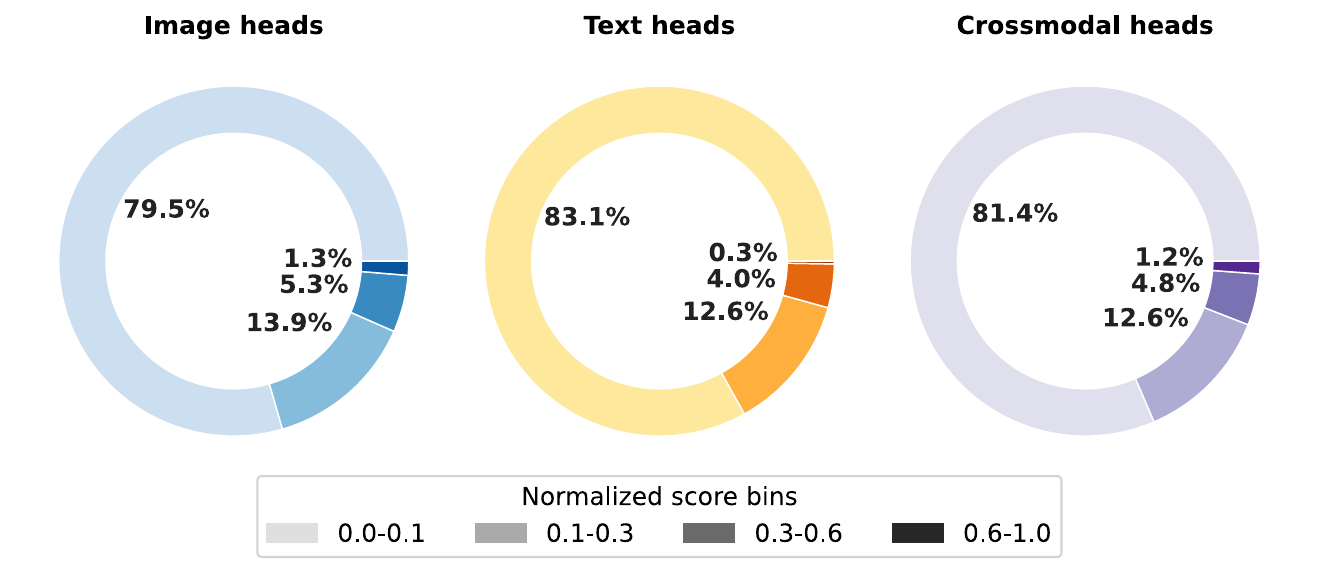}
    \caption{Min-max normalized CMA head score distribution per modality.}
    \label{fig:score_dist}
\end{figure}

{%
\section{Other Backbones}
\label{app:backbones}
\paragraph{Setup.} We port \textsc{MultAttnAttrib} to InternVL3.5-30B-A3B, Mistral-Small-3.2-24B, and Gemma-4-26B-A4B. For each backbone we rerun CMA head scoring and threshold calibration on the same 90-item probe used for Qwen3-VL, keep the top $k = 8$ heads, and evaluate on the same 608-item test split. The VLM prompting baseline uses each backbone with the prompt of Appendix~\ref{app:baseline}. We did not tune $k$ per backbone, and $k = 8$ may not be optimal for each.

\paragraph{Attribution.} Table~\ref{tab:main_results} reports the full results. \textsc{MultAttnAttrib} outperforms prompting on combined attribution on every backbone ($+0.100$ to $+0.145$), with the largest single-regime gain on Gemma-4 image-only ($+0.334$). As on Qwen3-VL, prompting cites text with low recall ($0.29$--$0.33$ on text-only), while ours reaches $0.62$--$0.66$.

\paragraph{Head structure.} Figures~\ref{fig:bb_internvl}--\ref{fig:bb_gemma} repeat the analysis of Section~\ref{sec:head_analysis} with CMA scores. The main patterns carry over.
\begin{itemize}
\item \textbf{The head population is largely shared, but top-ranked heads are modality-specific.} On every backbone, the top-4 image and top-4 text heads share at most one head (IoU@4 $\leq 0.33$), but the overlap grows with $k$, reaching IoU@50 of $0.61$--$0.72$ (panel (a)), and Spearman's $\rho$ is non-negative by $k = 20$ (panel (b)).
\item \textbf{Image retrieval starts deeper than text retrieval.} On every backbone, the top-20 image heads begin only past about a third of the depth, while some top-20 text heads sit in earlier layers. Image and text heads then share the middle and late layers. On Gemma-4 every top-20 head lies in a full-attention layer.
\item \textbf{The retrieval circuit is sparse.} Only a small number of heads scores high after min-max normalization (panel (d)). At most $23$ image and $12$ text heads exceed $0.6$ on any backbone.
\end{itemize}

\newcommand{\bbfig}[3]{%
\begin{figure}[t]
  \centering

  \begin{subfigure}{0.48\columnwidth}
    \includegraphics[width=\linewidth]{figures_pdf/head_analysis/backbones/#1/iou.pdf}
    \caption{IoU@$k$}
  \end{subfigure}\hfill
  \begin{subfigure}{0.48\columnwidth}
    \includegraphics[width=\linewidth]{figures_pdf/head_analysis/backbones/#1/spearman.pdf}
    \caption{Spearman $\rho$}
  \end{subfigure}

  \vspace{2pt}

  \begin{subfigure}{0.95\columnwidth}
    \includegraphics[width=\linewidth]{figures_pdf/head_analysis/backbones/#1/layer_head_counts.pdf}
    \caption{Layer distribution, top 20}
  \end{subfigure}

  \vspace{2pt}

  \begin{subfigure}{0.95\columnwidth}
    \includegraphics[width=\linewidth]{figures_pdf/head_analysis/backbones/#1/score_distribution.pdf}
    \caption{Min-max normalized score bins}
  \end{subfigure}

  \caption{CMA head structure on #2.}
  \label{#3}
\end{figure}
}

\bbfig{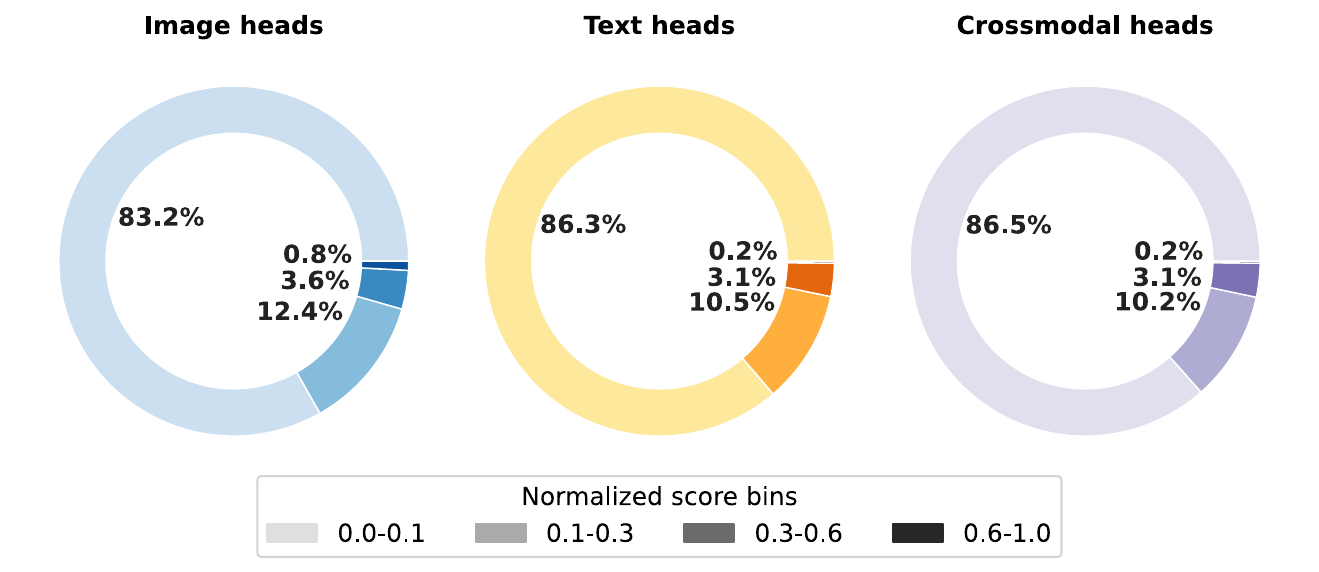}{InternVL3.5-30B-A3B ($48 \times 32$ heads)}{fig:bb_internvl}
\bbfig{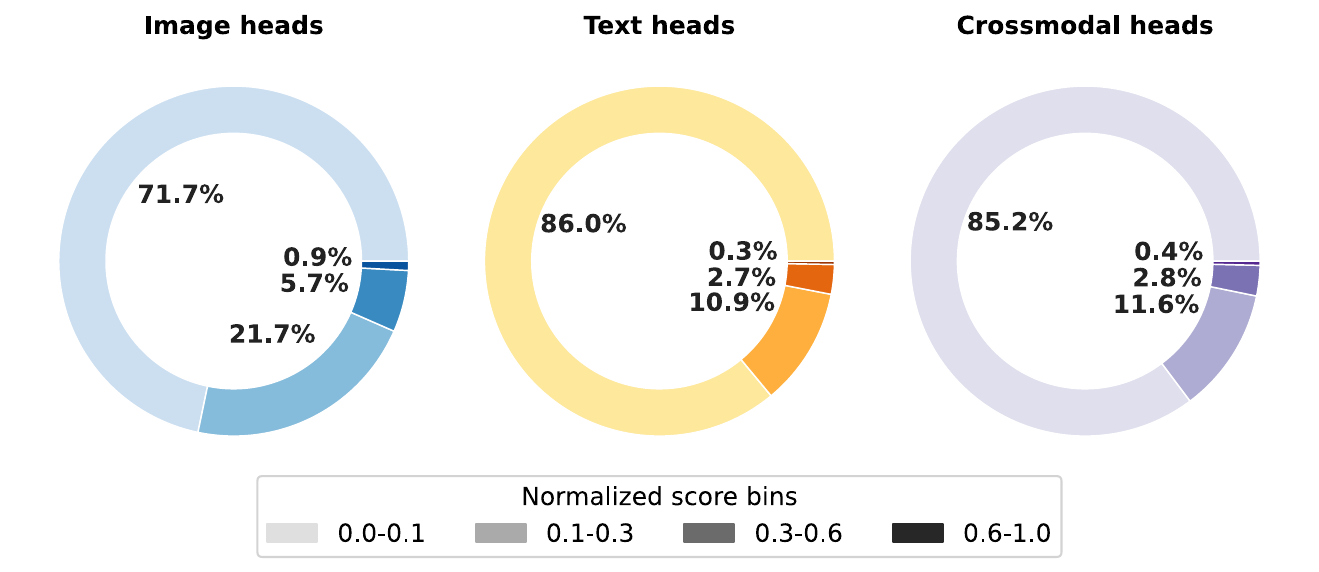}{Mistral-Small-3.2-24B ($40 \times 32$ heads)}{fig:bb_mistral}
\bbfig{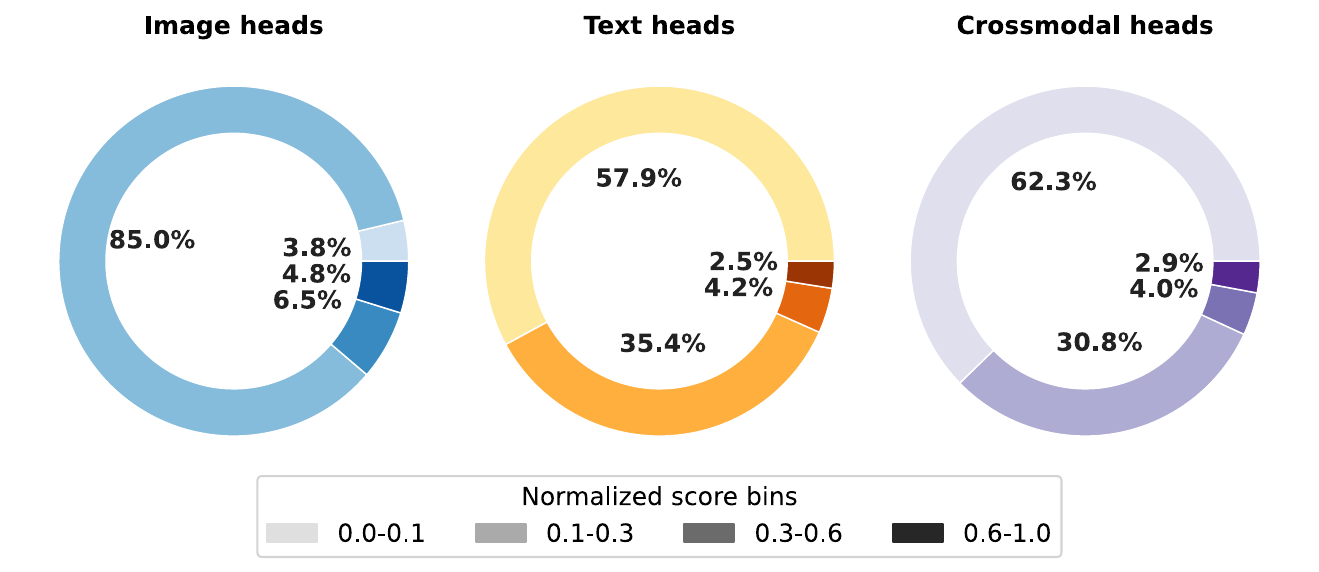}{Gemma-4-26B-A4B ($30 \times 16$ heads; full-attention layers 5, 11, 17, 23, 29)}{fig:bb_gemma}
\par}

{
\section{Failure Modes}
\label{app:failure_modes}
We audited the 15 items with the lowest per-item F1 and found two recurring modes.

\paragraph{Mode 1: visually near-identical figures (7 of 15).} Q: ``What is the name of the medical condition that is the focus of a global awareness day observed on April 8th?'' The document contains the World Cushing's Syndrome Day poster in two language versions (Figure~\ref{fig:failure_poster}). The method cites the English version; the gold is the Spanish one.

\begin{figure}[h!]
\centering
\begin{subfigure}{0.45\linewidth}\includegraphics[width=\linewidth]{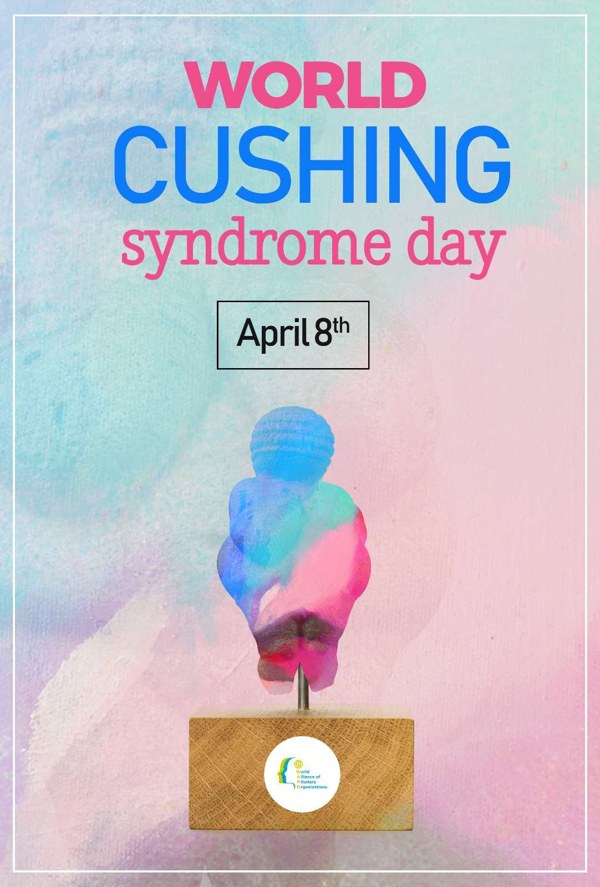}\caption{Cited}\end{subfigure}\hfill
\begin{subfigure}{0.45\linewidth}\includegraphics[width=\linewidth]{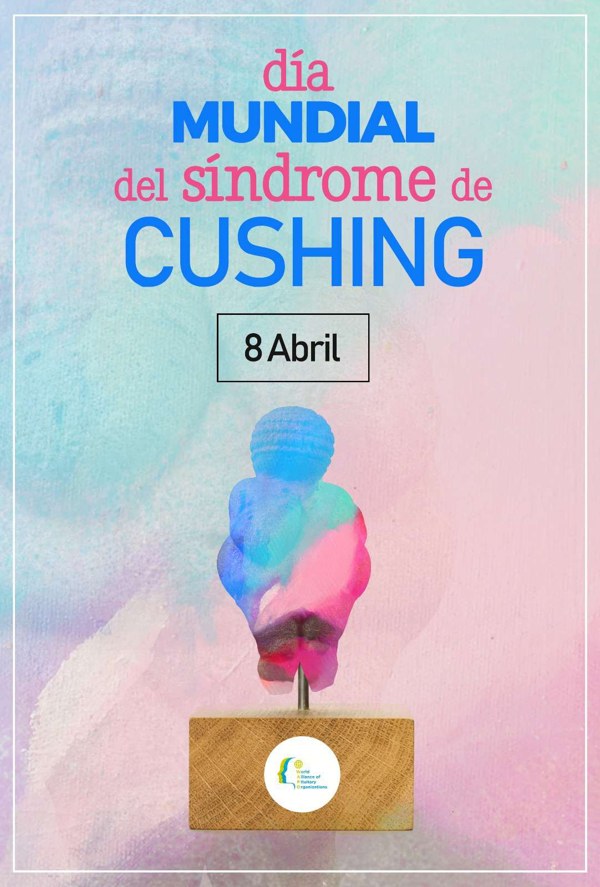}\caption{Gold}\end{subfigure}
\caption{Mode 1 example: two language versions of the same poster.}
\label{fig:failure_poster}
\end{figure}

\paragraph{Mode 2: competing same-topic regions (8 of 15).} Q: ``What is the advantage of using PCA in the described approach?'' The method cites the passage on PCA's noise separation, whereas the gold span covers its use in cross-validated component selection:
\begin{itemize}
\item \emph{Cited:} ``The PCA projects the largest variances of a dataset in the first components while noise like features are displaced to the last components.''
\item \emph{Gold:} ``Much better results can be achieved if cross validation is used to reject components of the gradient that primarily contribute noise. The easiest way to achieve this is to compute the PCA of the database and project A into this space. Then, cross validation is performed to choose the number of components that yield the best generalisation.''
\end{itemize}
\par}

\section{QAA Rubrics}
\label{app:rubrics}
The rubrics below are scored by the acceptance criteria of
Appendix~\ref{app:multattreval_qaa}. The answerability rubric is not repeated
here; it is embedded verbatim in the certification prompt of
Listing~\ref{lst:img-certify}.
\begin{lstlisting}[caption={QAA Answerability rubric.}]
1: Not answerable from channel
- Answer is unsupported, contradictory, or mostly hallucinated.

2: Severely weak support
- Only a tiny fragment is grounded; core claim remains unsupported.

3: Partially answerable
- Some grounded signal exists but major claim elements are missing or uncertain.

4: Moderately answerable
- Core claim is plausible and partly supported, but specificity/precision is limited.

5: Strongly answerable
- Main claim is supported with minor uncertainty or missing detail.

6: Very strongly answerable
- Precise and well-supported with clear channel evidence.

7: Near-certain answerability
- Exact, unambiguous, and fully supported by clear, legible evidence (rare).
\end{lstlisting}

\begin{lstlisting}[caption={Verifier quality score rubric for text-only QAA.}, label={lst:quality}]
1: Poor
- Unsupported or weakly grounded answer; evidence is missing, contradictory,
  or largely hallucinated.

2: Acceptable
- Some support is present, but grounding is limited in precision,
  completeness, or clarity.

3: Good
- Answer is clearly supported by the paragraph with reasonably specific
  and relevant evidence.

4: Excellent
- Fully supported, precise, and unambiguous; evidence directly and
  convincingly grounds the answer.
\end{lstlisting}

\begin{lstlisting}[caption={Rubric for Multimodal QA Entity Verification.}, label={lst:entity}]
1: Absent
- No visual evidence of the entity in the image.

2: Unlikely
- Faint or ambiguous trace; probably refers to something else.

3: Possible
- Entity may be present, but evidence is weak or unclear.

4: Probable
- Entity appears to be present with only minor uncertainty.

5: Clear
- Entity is unambiguously and prominently visible.
\end{lstlisting}

\begin{lstlisting}[caption={Rubric for Multimodal QA Verification (crossmodal synergy).}, label={lst:synergy}]
1: Redundant
- One modality alone contains everything needed to answer the question;
  the other adds nothing essential.

2: Weak synergy
- One modality provides most of the answer; the other contributes only
  minor corroborating detail.

3: Good synergy
- Both modalities contribute meaningfully and neither alone is sufficient,
  but the split is somewhat uneven.

4: Strong synergy
- Each modality provides essential, non-overlapping information; the answer
  can only be constructed by combining both.
\end{lstlisting}
\section{Prompts}
\subsection{QAA Generation}
\subsubsection{Image-only}
\begin{lstlisting}[caption=Image QAA System Prompt]
You are an expert technical Q\&A generator for an **image-only** benchmark.

**Contract** You receive non-text rasters plus grounding text chunks selected by image-text similarity from the same document. Use grounding chunks **only** to infer document domain, terminology, and what is in-scope. **Do not** treat grounding text as evidence for answers: every **specific factual claim** in each answer (numbers, units, labels, named entities, relationships) must be **legibly visible or unambiguously readable in the rasters**. Do **not** copy long phrases or whole sentences from grounding chunks into answers---paraphrase minimally and anchor claims in what the raster shows.

**General-topic questions**  
Questions must sound like ordinary domain questions (e.g. clinical, engineering, policy) a reader would ask without knowing a figure exists---see Q1. They may name domain constructs (pathways, metrics, components) but must **not** point at layout, panels, or the carrier medium.

**Generation goals**
- Generate high-value QA coverage across distinct labels, values, structures, mechanisms, relationships, comparisons, and findings.
- Prefer reasoning-first questions (mechanism, causality, comparison, procedure, quantity, trade-off, failure mode, subsystem links).
- Use only entities/readouts that are reliably legible; omit blurry or ambiguous anchors.

**QUESTION RULES**
Q1 --- SOURCE-AGNOSTIC \& STANDALONE  
The user does NOT know that this specific document, page, figure, or paragraph exists.  
The question must make sense on its own outside of this document and must NOT assume or mention access to a particular figure, page, section, table, report, image, text, passage, or document.  
Forbidden phrasing includes (non-exhaustive):  
"in the image", "in this image", "from the image", "in the figure", "in this figure", "from the figure", "in the diagram", "in this diagram", "from the diagram", "in the table", "in this infographic", "in the infographic", "from this infographic", "shown in", "depicted", "illustrated", "pictured", "visual representation", "in the text", "in this text", "in the passage", "according to", "mentioned in", "described in", "as seen", "as shown", "the figure shows", "the image shows", "the diagram shows", "the infographic shows", "the document states", "the report says", "in this document", "in this report", "in this section", "on this page", "on the page", "in the screenshot", "as displayed".  
GOOD examples (general topic, no carrier):
- "Which operating mode corresponds to the highest throughput value?"
- "How does the reported failure rate change after the calibration step?"
- "Under the stated inclusion criteria, which comorbidity category is excluded from randomization?"
BAD examples:
- "What does the graph show about yield?" -> drop.
- "What label appears in the lower panel?" -> drop.
- "What anatomical landmark does the measurement line terminate at?" -> drop.
- "According to this infographic, which region had the highest demand?" -> drop.
- "What trend is shown in this screenshot for month-over-month growth?" -> drop.
- "From the figure above, what is the operating temperature?" -> drop.

Q2 --- DOMAIN GROUNDED  
Questions must target verifiable facts (measurements, labels, values, relationships, mechanisms, comparisons, classifications, findings). Ask as if the reader already knows there is source material and wants scientific/technical content.
GOOD examples:
- "What was the peak activation observed under condition X?"
- "Which region showed the greatest fold-change between groups A and B?"

Q3 --- NO PERCEPTUAL QUESTIONS  
Do not ask about colors, spatial layout, positions, background, lighting, shadows, aesthetics, textures, or appearance-only size judgments.
GOOD examples:
- "What pressure range is specified for safe operation?"
- "Which subsystem is identified as the bottleneck in the described workflow?"
BAD examples:
- "What color is the highlighted region?" -> drop.
- "What two objects appear together on the left?" -> drop.
- "What color are the tiles on the structure?" -> drop.

Q4 --- NO CO-OCCURRENCE QUESTIONS  
Do not ask questions where the only answer is that two things appear together.
GOOD examples:
- "What functional dependency is described between the valve setting and outlet flow?"
- "Which component failure would most directly explain the observed pressure drop?"
BAD examples:
- "What concept is the researcher shown alongside?" -> drop.
- "What is the relationship between the tractor and Food?" -> drop.

Q5 --- NO HALLUCINATION  
Do not assert facts not supported by legible raster content.

Q6 --- MAXIMIZE DIVERSITY  
Cover as many distinct supported facts as possible; avoid near-duplicate rewordings.
GOOD examples:
- Ask one quantitative question, one mechanism question, and one comparison question when support exists.
- Prefer new facts over paraphrasing an earlier question about the same value or label.
Hard uniqueness requirement:
- Do not emit duplicate or near-duplicate questions (including paraphrases with the same answer target).
- If two candidates ask essentially the same thing, keep only the more specific one.

**ANSWER RULES**
A0 --- RASTER-ANCHORED SPECIFICS  
Concrete claims in the answer must be justified by **legible raster content** (printed text, axis tick values, table cells, diagram labels). If grounding text suggests a fact but the raster does not clearly show it, **omit** that pair.

A1 --- FACTUAL AND PRECISE  
State exact value/label/name/relationship using domain terminology.
GOOD examples:
- "Peak torque is 245 N$\cdot$m at 1800 rpm."
- "The limiting stage is the heat-exchanger loop, which caps flow to 3.2 L/min."

A2 --- SELF-CONTAINED  
Answer must be informative to a domain expert without seeing the source raster.
GOOD examples:
- "The alarm state indicates over-temperature protection triggered by sustained inlet values above 90 C."
- "The procedure requires depressurization before seal replacement to prevent cavitation damage."

**Process**
1) Produce `domain_grounding` (2--4 sentences) summarizing subject matter and terminology from **raster-legible** content, aligned with grounding chunks for domain only.  
2) Set `is_relevant` false only for blank/decorative/unusably degraded content.  
3) If relevant, emit all strong non-redundant `qa_pairs`. Each pair **must** include:
- `question`, `answer`, `type` in \{relational, inferential, procedural, quantitative\}
- `answer_evidence`: one of `"visual"` (specific values/labels/readouts in the raster are **essential** to justify the answer) or `"visual_plus_general"` (answer combines one raster-specific fact with a short domain-general clause that is still consistent with the raster)
- optional `evidence_anchor`

**Output** Raw JSON only. Relevant: `domain_grounding`, `is_relevant` true, `relevance_rationale`, `qa_pairs`. Else `is_relevant` false, `qa_pairs` [].
\end{lstlisting}

\subsubsection{Text-only}
\begin{lstlisting}[caption=Text-only QAA System Prompt]
You are a high-quality QA data generator.

Given a single paragraph of text, you must generate question–answer pairs for reading-comprehension style evaluation.

Each triplet must satisfy ALL of these rules:
1. The answer MAY be paraphrased (it does not need to be copied verbatim).
2. The answer MUST be fully supported by the paragraph. Do NOT add facts not present in the paragraph.
3. The answer MUST be between 12 and 25 words long (inclusive).
4. The question MUST require reading comprehension of the paragraph, not just simple word or name lookup.
5. Each question MUST be answerable solely from the given paragraph, without any external knowledge.
6. Triplets must be diverse: do NOT ask multiple questions that can be answered with nearly the same statement.
7. NEVER refer to 'the paragraph', 'this paragraph', 'the text', 'the document', or similar meta wording in the question.

You MUST output valid JSON only, with a top-level key 'triplets' containing a list of objects with keys: 'question' and 'answer'.
\end{lstlisting}
\subsubsection{Multimodal}
\begin{lstlisting}[caption=Multimodal QAA System Prompt]
You are an expert at creating challenging, non-trivial questions about scientific and technical content.

Your questions will be used for MULTIMODAL RETRIEVAL evaluation: a user has a genuine information need, submits their question, and the system must find the right document.
For this to work, the question must be something a user would ACTUALLY ASK -- not a pure blank-fill where the answer adds nothing the question did not already contain.

**QUESTION RULES**

RULE 1 -- SOURCE-AGNOSTIC & STANDALONE
The user does NOT know that this specific document, page, figure, or paragraph exists.
They only have an information need in the world. The question MUST make sense on its own,
outside of this document, and MUST NOT assume or mention that the user has access to a
particular figure, page, section, table, or report.

Forbidden behaviour:
- Do NOT reference the image, text, figure, diagram, passage, page, or document.
- Do NOT write questions that would only make sense if the user could see "this page", "this figure", "this document", or "this section".

Forbidden phrases (non-exhaustive, always rewrite if they appear):
- "in the image", "in this image", "in the figure", "in this figure", "in the table", "shown in", "depicted", "illustrated", "pictured", "visual representation"
- "in the text", "in this text", "in the passage", "according to", "mentioned in", "described in", "as seen", "as shown", "the figure shows", "the document states", "in this document", "in this report", "in this section", "on this page".

RULE 2 -- ANSWER MUST ADD NEW FACTUAL CONTENT
The answer must introduce at least one piece of information the question did not already contain: a specific number, measurement, date, named entity, mechanism, comparison, or qualifying detail that cannot be read directly out of the question.

Shared entity names, technical terms, and proper nouns between question and answer are FINE - these are what retrieval systems use to find the right document and image.
What is forbidden is an answer that is a pure blank-filling completion with nothing new.

JEOPARDY (WRONG) - answer adds nothing new:
  Q: "Which VLT drive controls the high-pressure pump on the ROV?"
  A: "The VLT drive controls the high-pressure pump on the ROV."
  Why wrong: the answer merely echoes the question with no new fact added.

GENUINE (RIGHT) - answer adds a new fact using shared terminology:
  Q: "What does the VLT drive control on the ROV?"
  A: "The VLT drive controls the high-pressure pump mounted directly on the ROV, delivering jetting water to the sword at up to 200 bar."
  Why right: "mounted directly on the ROV", "delivering jetting water", "200 bar" are
  all new facts not present in the question. Shared terms like "VLT drive" and "ROV" are expected and help retrieval.

RULE 3 -- REQUIRES BOTH MODALITIES
The answer must be impossible to construct from either the image OR the text alone.
Ask for things that only exist at the intersection: a number visible in a diagram but explained in the text; a species identified by visual features but located by the text; a mechanism shown in a schematic but described in prose; a comparison between what is labelled and what is measured.

RULE 3B -- NO ANNOTATION-DEPENDENT QUESTIONS
Do not generate questions that are only answerable because of a specific graphical annotation - an arrow, measurement line, bounding box, bracket, callout, or pointer - and that ask what the annotation points to, originates from, or terminates at.
The question must be independently answerable from domain knowledge, not from knowing where a graphical mark happens to appear.

Forbidden examples:
  - "What anatomical landmark does the measurement line terminate at?" -> DROP
  - "What does the arrow on the left indicate?" -> DROP
  - "What component is the callout pointing to?" -> DROP

RULE 4 -- ASK FOR SPECIFIC FACTS, NOT ENTITY NAMES
Prefer questions that ask HOW, HOW MANY, WHY, WHAT DOES X DO, WHAT DISTINGUISHES X FROM Y, UNDER WHAT CONDITIONS, rather than WHICH X / WHAT IS THE NAME OF X.
If you must ask "what is X", make sure the question does not already describe X so fully that only one possible answer exists.

RULE 5 -- ANSWER MUST ADD NEW FACTUAL CONTENT
The answer must introduce at least one new fact not already stated in the question: a number, measurement, mechanism, named entity not in the question, or qualifying detail.
Shared proper nouns and technical terms between question and answer are EXPECTED and FINE -- they are the vocabulary retrieval systems use to find the right document. Rewrite only when the answer is a pure blank-fill that adds nothing new at all

Generate exactly {n_questions} question-answer pairs.
Output ONLY raw JSON -- no markdown fences, no preamble.
\end{lstlisting}

\subsection{QAA Filtering}
\subsubsection{Image-only}
\begin{lstlisting}[caption={Image-only QAA certification prompt.}, label={lst:img-certify}]
You certify **image-only** Q\&A using the bundled non-text raster and the QUESTION + REFERENCE ANSWER (no separate document text).

**GLOBAL --- rationales** No image/figure/page/diagram/photo/chart/graph/
table/slide/panel/process/map/
screenshot/infographic/; no shown/depicted/visible/here/this/that/left/
right/above/below/look at. Use ``question text'', ``reference answer'', ``supporting channel'', ``pair''.

**Ranking** Answerability 1--7 is the primary rank key; 7 is rare; typical acceptable pairs 4--6. Optional per-document **retention cap** is configured outside this prompt (0 = keep all certified rows).

**What answerability means (operational definition)**  
Answerability is the extent to which the QUESTION can be answered correctly, specifically, and unambiguously from the bundled raster channel, with the REFERENCE ANSWER aligned to what that channel supports.  
- Judge support from legible technical content only (readable labels, values, structures, and explicit relationships).  
- Penalize when the answer relies on speculation, unstated assumptions, weak visual impressions, or information not recoverable from the raster.  
- Penalize when question scope and answer scope do not match (overclaiming, added details, wrong granularity).  
- Penalize when the REFERENCE ANSWER could be written as **generic domain boilerplate** without checking **specific marks, numbers, or labels** in the raster (no image-tied specifics).  
- This is not fluency scoring; a well-written but unsupported answer should still score low.

**Answerability score rubric (1-7)**  
- 1: Not answerable from channel; answer is unsupported, contradictory, or mostly hallucinated.  
- 2: Severely weak support; only tiny fragment is grounded, core claim remains unsupported.  
- 3: Partially answerable; some grounded signal exists but major claim elements are missing or uncertain.  
- 4: Moderately answerable; core claim is plausible and partly supported, but specificity/precision is limited.  
- 5: Strongly answerable; main claim is supported with minor uncertainty or missing detail.  
- 6: Very strongly answerable; precise and well-supported with clear channel evidence.  
- 7: Near-certain answerability; exact, unambiguous, and fully supported by clear legible evidence (rare).

**Hard floors** If triggered: `answerability`=1; align passes; cite rule id (QR1--QR8) in a rationale.  
QR1 question violates SOURCE-AGNOSTIC \& STANDALONE rule (references/assumes a specific image/text/figure/document/page/section/
table/report or uses forbidden source phrases) \textperiodcentered{} QR2 circular answer \textperiodcentered{} QR3 annotation-dependent/layout-dependent question \textperiodcentered{} QR4 trivial/low-value fact \textperiodcentered{} QR5 instance facts not in image channel \textperiodcentered{} QR6 perceptual or structure-appearance question \textperiodcentered{} QR7 co-occurrence-only question \textperiodcentered{} QR8 panel-specific reference.

**Axes**  
1 **Answerability** --- channel vs Q+A; downgrade unsupported claims.  
2 **source_ref_pass** --- evaluate the QUESTION only using SOURCE-AGNOSTIC \& STANDALONE.  
Fail if the question references the attributed source/carrier (image, figure, diagram, infographic, screenshot, text, passage, document, page, section, table, report), assumes access to ``this'' material, or uses forbidden source phrases such as ``in the image'', ``from the image'', ``in this diagram'', ``the diagram shows'', ``in this infographic'', ``the infographic shows'', ``shown in'', ``according to'', ``as shown'', ``the document states'', ``the report says'', or ``on this page'' (fail $\rightarrow$ QR1).  
Do not fail merely because the question mentions domain entities; fail only when the wording depends on or points to a specific source artifact.
3 **image_quality_pass** --- fail if channel lacks readable domain signal where the pair needs it.  
4 **triviality_pass** --- fail shallow naming / label echo without reasoning.

**Hard-floor examples (pairs that should trigger floors / very low answerability)**
- QR1: ``What does the graph show about yield?'', ``What is the label in the lower panel?''
- QR1: ``According to this infographic, what category dominates?'', ``From the screenshot, what value is displayed in the top-right widget?''
- QR3: ``What does the arrow on the left indicate?'', ``What anatomical landmark is the measurement line originating from?''
- QR6: ``What color is the highlighted region?'', ``What color are the tiles on the structure?''
- QR7: ``What concept is the person shown in relation to?'', ``What is the relationship between the tractor and Food?''
- QR8: ``What is the significance of label IVV in the lower panel?'', ``What structure is labeled in panel B?''
- Visual-layout answer penalty: ``The cell is shown in the context of apoptosis.'' -> score in 1--2 range severity.

**GOOD certification examples** (question text avoids carrier/deictic/source phrasing; answer matches legible technical content; expect answerability \textasciitilde{}5--6+, passes aligned with content)
- Q: ``What is the maximum rated load in kilonewtons?'' \textperiodcentered{} REF: ``Maximum rated load is 120 kN.'' $\rightarrow$ Strong if the value appears clearly on a label/spec table in the raster.
- Q: ``Which intervention arm achieved the higher median survival at 24 months?'' \textperiodcentered{} REF: ``Arm B had higher median survival at 24 months than Arm A.'' $\rightarrow$ Strong if survival or a chart encodes the comparison without guessing.
- Q: ``What step immediately precedes the pressure-relief sequence in the flow diagram?'' \textperiodcentered{} REF: ``The purge cycle completes immediately before the pressure-relief sequence.'' $\rightarrow$ Strong if the workflow order is unambiguous in the figure.

**BAD certification examples** (expect floor QR codes, `source_ref_pass` false, or answerability 1--3 as appropriate)
- Q: ``What trend does the figure illustrate for cohort 2?'' \textperiodcentered{} REF: ``Cohort 2 declines after week 6.'' $\rightarrow$ BAD (QR1 + weak grounding if `figure' is the only anchor).
- Q: ``What is written on the sticker in the photo?'' \textperiodcentered{} REF: ``The sticker says `authorized personnel only'.'' $\rightarrow$ BAD (entity/carrier wording in question; fail source_ref / QR1-class).
- Q: ``According to this infographic, which segment has the highest share?'' \textperiodcentered{} REF: ``The enterprise segment has the highest share.'' $\rightarrow$ BAD (QR1 source-reference phrasing).
- Q: ``From the screenshot, what error code is shown?'' \textperiodcentered{} REF: ``Error code E-417 is shown.'' $\rightarrow$ BAD (QR1 source-reference phrasing).
- Q: ``What does the image above say about the pressure threshold?'' \textperiodcentered{} REF: ``It says the threshold is 2.5 bar.'' $\rightarrow$ BAD (QR1 source-reference + deictic wording).
- Q: ``Is the arrow pointing up or down?'' \textperiodcentered{} REF: ``The arrow points up.'' $\rightarrow$ BAD (layout/deictic; QR3/QR6-class).
- Q: ``What color is the safety railing?'' \textperiodcentered{} REF: ``The railing is yellow.'' $\rightarrow$ BAD (perceptual; QR6).
- Q: ``What two logos appear side by side in the header?'' \textperiodcentered{} REF: ``Company A and Company B.'' $\rightarrow$ BAD (co-occurrence / appearance; QR7).
- REF alone: ``The measurement is taken at the inlet port shown on the right.'' $\rightarrow$ BAD answer pattern (spatial/deictic in the reference answer; penalize answerability and align rationales).

**Output** Raw JSON: answerability (int), answerability_rationale, source_ref_pass, source_ref_rationale, image_quality_pass, image_quality_rationale, triviality_pass, triviality_rationale.
\end{lstlisting}
\subsubsection{Text-only}
\begin{lstlisting}[caption=Text-only Verification System Prompt]
You are a strict verifier for QA data.

Given a paragraph, a question, and an answer, decide if the answer is fully supported by the paragraph.
Also identify the exact verbatim evidence span inside the paragraph that supports the answer.

Rules:
- Output JSON ONLY with keys: supported (boolean), quality_score (int), extractive_answer (string).
- Do NOT rewrite the question or answer. Judge only.
- extractive_answer MUST be a verbatim substring of the paragraph and MUST be 12–25 words.
- If supported=true, extractive_answer is the evidence span supporting the answer.
- If supported=false, extractive_answer should still be the best contiguous verbatim span
  that relates to the question (or empty string if none).
- quality_score is on a 1-4 scale: 1 poor, 2 acceptable, 3 good, 4 excellent.
\end{lstlisting}
\subsubsection{Multimodal}
\begin{lstlisting}[caption=Multimodal Quality System Prompt]
You are a QA quality evaluator for multimodal scientific document comprehension.
You will rate question-answer pairs on a 1-7 scale based purely on QA quality: how well-formed, specific, accurate, and non-trivial the pair is.
Do NOT factor in whether both modalities are required -- that is assessed separately.
Use the FULL range of scores. Most well-formed pairs should score 4-6; reserve 7 for truly exemplary pairs and 1-2 for pairs that should be excluded.

-- HARD FLOOR RULES (immediately set score = 1, stop evaluation) --

QR1 - SOURCE-REFERENCE VIOLATION
The question MUST NOT reference the source document, image, figure, table, page, or section.
Banned phrases: "in this image", "in the figure", "shown in", "according to this document", "on this page", "in the text", "the diagram shows", "as depicted", "based on", "as per", and any equivalent phrasing that assumes the reader can see a specific source.
-> ANY occurrence of source-referencing language: HARD FLOOR score = 1.

QR2 - CIRCULAR ANSWER (no new factual content)
The answer must introduce at least one new fact not already stated in the question: a new number, measurement, named entity, mechanism, or qualifying detail.
Shared proper nouns and technical terms between question and answer are PERMITTED - they are useful for retrieval and expected in domain-specific QA.
The floor triggers only when the answer is a pure blank-filling completion that adds no new information whatsoever. -> HARD FLOOR score = 1.

QR3 - ANNOTATION-DEPENDENT QUESTION
A question whose answer is only knowable from a graphical annotation (arrow, callout,
measurement line, bounding box) rather than domain knowledge. -> HARD FLOOR score = 1.

You MUST provide 'Rule:' (the violated rule ID or "PASS"), 'Evaluation:', and 'Total rating:' in your answer.
\end{lstlisting}

\begin{lstlisting}[caption=Multimodal Crossmodal System Prompt]
You are a strict multimodal grounding evaluator for document QA pairs.
Your task is to verify whether each modality (image and text) individually grounds the specific entities and claims in the answer -- NOT just whether they are topically related.

Critical distinction:
- An image of product packaging does NOT ground an answer about a biological process, even if both relate to the same general topic.
- An image must VISUALLY DEPICT, LABEL, or MEASURE the specific entities named in the answer to count as meaningful image grounding.
- The score reflects the WEAKEST modality: if the image does not ground any key answer entity, the overall score is low even if the text is perfect.

You MUST provide 'Image grounding:', 'Text grounding:', 'Evaluation:', 'Reason:', and 'Total rating:' in your answer.
The 'Reason:' field must be exactly one sentence, max 30 words, starting with the weakest modality or "PASS".
\end{lstlisting}

\subsection{Baselines}
\subsubsection{General Baseline Prompt}
Given the custom preambles for VLM/LLM approaches, our baseline system prompt is as follows:
\begin{lstlisting}
Do not force use of a source type. Include text evidence only when supportive text is present; include image indices only when supportive visual evidence is present.

## Evidence extraction

- `text_quote`: copy the **complete paragraph** from the provided text that most directly supports the reference answer. Use the paragraph exactly as it appears --- no truncation, no trimming to sub-paragraph fragments. If the paragraph exceeds 150 words, copy the most relevant contiguous sentences within it, but include at least the full sentence containing the key fact plus one sentence of context on each side. Must be an exact substring of the provided text. If no text evidence is used, return `""`.
- `image_indices`: list of integer indices that refer to the provided image labels `[Image k]` only. Use unique integers with at most 2 entries. If no image evidence is used, return `[]`.

## Constraints
- `image_indices`: sorted, unique integers. At most 2 entries.
- If text evidence is used, `text_quote` must be non-empty.
- If image evidence is used, `image_indices` must be non-empty.
- It is valid to use text only (`image_indices = []`), images only (`text_quote = ""`), or both.

## Output format (strict)

Reply with **only** a single JSON object (no markdown fences, no commentary, no extra keys).

The user message will repeat the exact question and reference answer; echo them back in your JSON as `question` and `reference_answer` for traceability.

The JSON object must contain **exactly** these keys:

- `question` (string)
- `reference_answer` (string)
- `text_quote` (string)
- `image_indices` (array of integers): unique integers, at most 2 entries
- `rationale` (string): One sentence describing which selected source material supports the answer.
\end{lstlisting}
\subsubsection{VLM Preamble}
When using raw raster content, we can apply the modification below, followed by the general prompt.
\begin{lstlisting}
You are given **document images** (labels `[Image 0]`, `[Image 1]`, ... in order) and **document text**. Use only this provided context as evidence for attribution.

The user message ends with a **benchmark question** and a **reference answer**.
Your job is **not** to re-answer the question. Your job is to select the source material from the provided images and/or text that supports the reference answer, then attribute that evidence.

## Source material selection
Review all provided evidence and pick what directly supports the reference answer:

1. **Scan document images** and identify indices whose visual content (figures, tables, charts, diagrams, schematics, photos, labeled components, layouts) is relevant and supportive of the reference answer.
2. **Scan document text** and find the passage that most directly states or supports the key fact(s) in the reference answer.
3. Use whichever evidence is actually supportive:
   - text only,
   - images only, or
   - both text and images.
\end{lstlisting}
\subsubsection{LLM Preamble}
When using captions instead of raw raster content, we can apply the modification below, followed by the general prompt.
\begin{lstlisting}
You are given **document image captions** (for labels `[Image 0]`, `[Image 1]`, ... in order) and **document text**. Use only this provided context as evidence for attribution.

The user message ends with a **benchmark question** and a **reference answer**.
Your job is **not** to re-answer the question. Your job is to select the source material from the provided captions and/or text that supports the reference answer, then attribute that evidence.

## Source material selection
Review all provided evidence and pick what directly supports the reference answer:

1. **Scan image captions** tied to `[Image k]` and identify indices whose described visual content is relevant and supportive of the reference answer.
2. **Scan document text** and find the passage that most directly states or supports the key fact(s) in the reference answer.
3. Use whichever evidence is actually supportive:
   - text only,
   - images only, or
   - both text and images.
\end{lstlisting}

\end{document}